\documentclass[10pt,journal,compsoc]{IEEEtran}

\ifCLASSOPTIONcompsoc
  \usepackage[nocompress]{cite}
\else
  \usepackage{cite}
\fi
\ifCLASSINFOpdf
  \usepackage[pdftex]{graphicx}
\else
\fi

\usepackage{epstopdf}
\usepackage{dashbox}
\usepackage{amsmath}
\usepackage{amssymb}
\usepackage{amsthm}
\usepackage{bm}
\usepackage{array}
\usepackage{threeparttable}

\ifCLASSOPTIONcompsoc
  \usepackage[caption=false,font=footnotesize,labelfont=sf,textfont=sf]{subfig}
\else
  \usepackage[caption=false,font=footnotesize]{subfig}
\fi

\usepackage{url}

\newcolumntype{P}[1]{>{\centering\arraybackslash}p{#1}}
\newcolumntype{M}[1]{>{\centering\arraybackslash}m{#1}}
\usepackage{multirow}

\begin{document}

\title{On the Reconstruction of Face Images from Deep Face Templates}

\author{Guangcan~Mai,
    Kai~Cao,
        Pong~C.~Yuen$^\ast$,~\IEEEmembership{Senior~Member,~IEEE,}
        and~Anil~K.~Jain,~\IEEEmembership{Life~Fellow,~IEEE}
\IEEEcompsocitemizethanks{
  \IEEEcompsocthanksitem Guangcan~Mai and Pong~C.~Yuen are with
  the Department of Computer Science, Hong Kong Baptist University, Hong Kong SAR, CHINA. \protect\\
  E-mail:\{csgcmai,pcyuen\}@comp.hkbu.edu.hk;
  \IEEEcompsocthanksitem Kai~Cao and Anil~K.~Jain are with the Department of Computer Science and Engineering, Michigan State University, MI, 48824, USA. \protect\\
  E-mail: \{kaicao,jain\}@cse.msu.edu}
  \thanks{$^\ast$ Corresponding author}\
}

\markboth
{Mai \MakeLowercase{\textit{et al.}}: On the Reconstruction of Face Images from Deep Face Templates}{}

\IEEEtitleabstractindextext{%
\begin{abstract}
  
State-of-the-art face recognition systems are based on deep (convolutional) neural networks.  
Therefore, it is imperative to determine to what extent face templates derived from deep networks can be inverted to obtain the original face image.
In this paper, we study the vulnerabilities of a state-of-the-art face recognition system based on template reconstruction attack.
We propose a neighborly de-convolutional neural network (\textit{NbNet}) to reconstruct face images from their deep templates.
In our experiments, we assumed that no knowledge about the target subject and the deep network are available.
To train the \textit{NbNet} reconstruction models, we augmented two benchmark face datasets (VGG-Face and Multi-PIE) with a large collection of images synthesized using a face generator.
The proposed reconstruction was evaluated using type-I (comparing the reconstructed images against the original face images used to generate the deep template) and type-II (comparing the reconstructed images against a different face image of the same subject) attacks. 
Given the images reconstructed from \textit{NbNets}, we show that 
for verification, we achieve TAR of 95.20\% (58.05\%) on LFW under type-I (type-II) attacks @ FAR of 0.1\%. 
Besides, 96.58\% (92.84\%) of the images reconstructed from templates of partition \textit{fa} (\textit{fb}) can be identified from partition \textit{fa} in color FERET.
Our study demonstrates the need to secure deep templates in face recognition systems.
\end{abstract}

}
\maketitle
\IEEEdisplaynontitleabstractindextext
\IEEEpeerreviewmaketitle

\IEEEraisesectionheading{\section{Introduction}
\label{sec:introduction}}

\IEEEPARstart{F}{ace}
recognition systems are being increasingly used for secure access in applications ranging from personal devices (e.g., iPhone X\footnote{\url{https://www.apple.com/iphone-x/#face-id}} and Samsung S8\footnote{\url{http://www.samsung.com/uk/smartphones/galaxy-s8/}}) to access control (e.g., banking\footnote{\url{https://goo.gl/6TGcrr}} and border control\footnote{\url{https://goo.gl/ViVdDY}}).
In critical applications, face recognition needs to meet stringent performance requirements, including low error rates and strong system security. 
In particular, the face recognition system must be resistant to spoofing (presentation) attacks and template invertivility.
Therefore, it is critical to evaluate the vulnerabilities of a face recognition system to these attacks and devise necessary countermeasures.
To this end, several attack mechanisms (such as hill climbing~\cite{adler2003sample,galbally2010vulnerability,feng2014masquerade}, spoofing~\cite{wen2015face,patel2016spoofdetect,siqiliu20163d,shao2017deep,liu2018learning}, and template reconstruction (template invertibility)~\cite{mignon2013reconstructing}) have been proposed to determine the vulnerabilities of face recognition systems.

\begin{figure}[t]
  \centering
  \includegraphics[width=.84\linewidth]{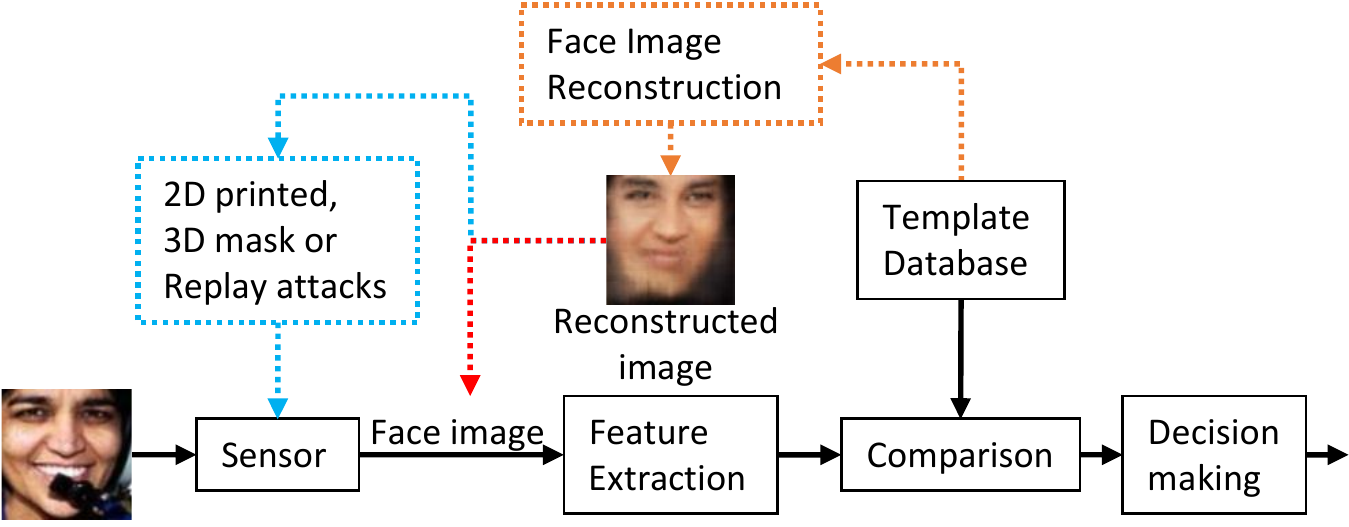}
  \caption{Face recognition system vulnerability to template reconstruction attacks. Face image of a target subject is reconstructed from the corresponding template to gain system access by (a) creating a fake face (for example, a 2D printed image or 3D mask) (blue box) or (b) injecting a reconstructed face image directly into the feature extractor (red box).
  }
  \label{fig:intro_template_rec}
\end{figure}

\begin{figure}[t]
  \centering
  \subfloat[Successful match]{
    \begin{minipage}{.44\linewidth}
    \centering
      {\includegraphics[width=.23\linewidth]{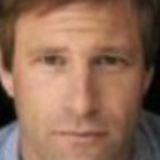}} 
      {\includegraphics[width=.23\linewidth]{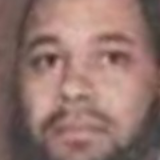}} 
      {\includegraphics[width=.23\linewidth]{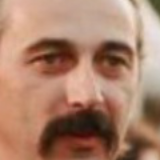}}
      {\includegraphics[width=.23\linewidth]{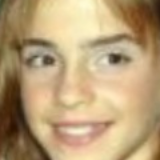}}\\
      \scriptsize{0.84\hspace{0.13\linewidth} 0.78\hspace{0.13\linewidth} 0.82\hspace{0.13\linewidth} 0.93}\vspace{0.1cm}\\
      {\includegraphics[width=.23\linewidth]{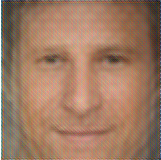}}\hspace{0.02cm}
      {\includegraphics[width=.23\linewidth]{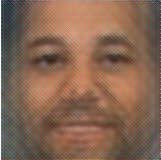}}\hspace{0.02cm}
      {\includegraphics[width=.23\linewidth]{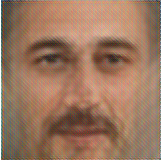}}\hspace{0.02cm}
      {\includegraphics[width=.23\linewidth]{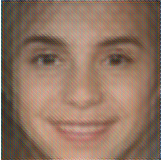}}
    \end{minipage}
      \label{fig:lfw_gd_rec}
     }
 	\hspace{.1cm}
    \subfloat[Unsuccessful match]{
    \begin{minipage}{.44\linewidth}
      \centering
      {\includegraphics[width=.23\linewidth]{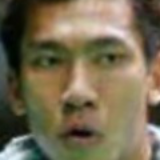}} 
      {\includegraphics[width=.23\linewidth]{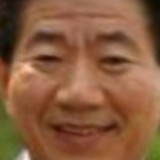}} 
      {\includegraphics[width=.23\linewidth]{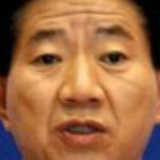}}
      {\includegraphics[width=.23\linewidth]{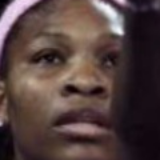}}\\
      \hspace{0.05\linewidth}\scriptsize{0.09\hspace{0.13\linewidth} 0.10\hspace{0.13\linewidth} 0.12\hspace{0.13\linewidth} 0.13}\vspace{0.1cm}\newline
      {\includegraphics[width=.23\linewidth]{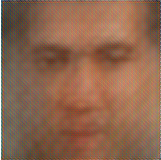}}\hspace{0.02cm}
      {\includegraphics[width=.23\linewidth]{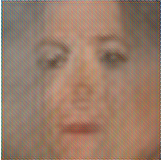}}\hspace{0.02cm}
      {\includegraphics[width=.23\linewidth]{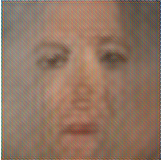}}\hspace{0.02cm}
      {\includegraphics[width=.23\linewidth]{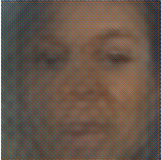}}
    \end{minipage}
      \label{fig:lfw_bd_rec}
     }  
    \caption{Example face images reconstructed from their templates using the proposed method (VGG-NbB-P). The top row shows the original images (from LFW) and the bottom row shows the corresponding reconstructions. The numerical value shown between the two images is the cosine similarity between the original and its reconstructed face image.
    The similarity threshold is 0.51 (0.38) at FAR = 0.1\% (1.0\%).}
    \label{fig:rec_example}
\end{figure}

In this paper, we focus on the vulnerability of a face recognition system to template invertibility or reconstruction attacks. 
In a template reconstruction attack (Fig.~\ref{fig:intro_template_rec}), we want to determine if face images can be successfully reconstructed from the face templates of target subjects and then used as input to the system to access privileges. 
Fig.~\ref{fig:rec_example} shows examples of face images reconstructed from their deep templates by the proposed method. 
Some of these reconstructions are successful in that they match well with the original images (Fig.~\ref{fig:rec_example}~(a)), while others are not successful (Fig.~\ref{fig:rec_example}~(b)). 
Template reconstruction attacks generally assume that templates of target subjects and the corresponding \textit{black-box} template extractor can be accessed.
These are reasonable assumptions because:
(a) templates of target users can be exposed in hacked databases\footnote{\url{https://goo.gl/QUMHpv}}$^,$\footnote{\url{https://goo.gl/KdxzqT}}, and 
(b) the corresponding \textit{black-box} template extractor can potentially be obtained by purchasing the face recognition SDK.
To our knowledge, almost all of the face recognition vendors store templates without template protection, while some of them protect templates with specific hardware (e.g., Secure Enclave on A11 of iPhone~X~\cite{apple2017face}, TrustZone on ARM\footnote{https://www.arm.com/products/security-on-arm/trustzone}). 
\begin{table*}[!t]
	\begin{threeparttable}
		\caption{Comparison of major algorithms for face image reconstruction from their corresponding templates}
		\centering
		\setlength{\extrarowheight}{.4mm}
		\begin{tabular*}{\linewidth}{M{0.11\linewidth}|M{0.13\linewidth}|M{0.46\linewidth}|M{0.2\linewidth}}
			\hline 
			\textbf{Algorithm} & \textbf{Template features} & \textbf{Evaluation} & \textbf{Remarks} \\ 
			\hline \hline 
			MDS~\cite{mohanty2007scores} & PCA, BIC, COTS &  Type-I attack\tnote{a}: TAR of 72\% using \textit{BIC}\tnote{b} and 73\% using \textit{COTS}\tnote{c} at an FAR of 1.0\% on FERET  & Linear model with limited capacity\\ 
			\hline 
			RBF regression~\cite{mignon2013reconstructing} & LQP~\cite{hussain2012face}  & Type-II attack\tnote{d}: 20\% rank-1 identification error rate on FERET; EER = 29\% on LFW;  & RBF model may have limited generative capacity\\ 
			\hline 
			CNN~\cite{zhmoginov2016inverting} & Final feature of FaceNet~\cite{Schroff_2015_CVPR} & \multirow{4}{\linewidth}{Reported results were mainly based on visualizations and no comparable statistical results was reported}  & \textbf{White-box} template extractor was assumed\\ \cline{1-2}\cline{4-4}
			Cole et. al., ~\cite{cole2017synthesizing} &{Intermediate feature of FaceNet~\cite{Schroff_2015_CVPR}}\tnote{e}  &  & High-quality images (e.g., front-facing, neutral-pose) are required for training.\\
			\hline
			This paper &Final feature of FaceNet~\cite{Schroff_2015_CVPR} & Type-I attack: TAR\tnote{f} of 95.20\% (LFW) and 73.76\% (FRGC v2.0) at an FAR of 0.1\%; rank-1 identification rate 95.57\% on color FERET \newline Type-II attack: TAR of 58.05\% (LFW) and 38.39\% (FRGC v2.0) at an FAR of 0.1\%; rank-1 identification rate 92.84\% on color FERET & Requires a large number of images for network training \\
			\hline 
		\end{tabular*}
		\begin{tablenotes}
			\item[a] {Type-I attack} refers to matching the reconstructed image against the face image from which the template was extracted.
			\item[b] \textit{BIC} refers to Bayesian intra/inter-person classifier~\cite{moghaddam1998beyond}.
			\item[c] \textit{COTS} refers to commercial off-the-shelf system. A local-feature-based COTS was used in~\cite{mohanty2007scores}.
			\item[d] {Type-II attack} refers to matching the reconstructed image against a face image of the same subject that was not used for template creation.
			\item[e] Output of 1024-D `avgpool' layer of the ``NN2'' architecture. 
			\item[f] TAR for LFW and FRGC v2.0 cannot be directly compared because their similarity thresholds differ.
		\end{tablenotes}
		\label{tb:facereview}
	\end{threeparttable}
\end{table*}
Note that unlike traditional passwords, biometric templates cannot be directly protected by standard ciphers such as AES and RSA since the matching of templates needs to allow small errors caused by intra-subject variations~\cite{nandakumar2015biometric,jain201650}.
Besides, state-of-the-art template protection schemes are still far from practical because of the severe trade-off between matching accuracy and system security~\cite{mai2016binary,feng2010hybrid}.

Face templates are typically compact binary or real-valued feature representations\footnote{As face templates refer to face representations stored in a face recognition system, these terms are used interchangeably in this paper.} that are extracted from face images to increase the efficiency and accuracy of similarity computation. 
Over the past couple of decades, a large number of approaches have been proposed for face representations.
These representations can be broadly categorized into (i) shallow~\cite{tur1991pca,ahonen2006lbp,hussain2012face}, and (ii) deep (convolutional neural network or CNN) representations~\cite{Parkhi15,Schroff_2015_CVPR,hayat2017joint}, according to the depth of their representational models\footnote{Some researchers~\cite{Parkhi15} refer to shallow representations as those that are not extracted using deep networks.}.
Deep representations have shown their superior performances in face evaluation benchmarks (such as LFW~\cite{learned2016labeled}, YouTube Faces~\cite{wolf2011face,Schroff_2015_CVPR}, and NIST IJB-A~\cite{klare2015pushing,hayat2017joint}).
Therefore, it is imperative to investigate the invertibility of deep templates to determine their vulnerability to template reconstruction attacks. 
However, to the best of our knowledge, no such work has been reported.

In our study of template reconstruction attacks, we made no assumptions about subjects used to train the target face recognition system. 
Therefore, only public domain face images were used to train our template reconstruction model.
The available algorithms for face image reconstruction from templates~\cite{mignon2013reconstructing,mohanty2007scores}\footnote{MDS method  in the context of template reconstructible was initially proposed for reconstructing templates by matching scores between the target subject and attacking queries. 
However, it can also be used for template reconstruction attacks~\cite{mohanty2007scores}.},\cite{zhmoginov2016inverting,cole2017synthesizing} are summarized in Table \ref{tb:facereview}.
The generalizability of the published template reconstruction attacks~\cite{mohanty2007scores,mignon2013reconstructing} is not known, as all of the training and testing images used in their evaluations were subsets of the same face dataset.
No statistical study in terms of template reconstruction attack has been reported in~\cite{zhmoginov2016inverting,cole2017synthesizing}.

To determine to what extent face templates derived from deep networks can be inverted to obtain the original face images, a reconstruction model with sufficient capacity is needed to invert the complex mapping used in the deep template extraction model~\cite{Goodfellow-et-al-2016}.
De-convolutional neural network (D-CNN)\footnote{Some researchers refer to D-CNNs as CNNs. However, given that its purpose is the inverse of a CNN, we distinguish D-CNN and CNN.}~\cite{zeiler2010deconvolutional,zeiler2011adaptive,gao2017pixel} is one of the straightforward deep models for reconstructing face images from deep templates. 
To design a D-CNN with sufficient model capacity\footnote{The ability of a model to fit a wide variety of functions \cite{Goodfellow-et-al-2016}. 
},  one could increase the number of output channels (filters) in each de-convolution layer \cite{zagoruyko2016wide}. 
However, this often introduces noisy and repeated channels since they are treated equally during the training. 

To address the issues of noisy (repeated) channels and insufficient channel details, inspired by \textit{DenseNet} ~\cite{huang2016densely} and \textit{MemNet}~\cite{tai2017memnet}, we propose a neighborly de-convolutional network framework (\textit{NbNet}) and its building block, neighborly de-convolution blocks (NbBlocks). 
The NbBlock produces the same number of channels as a de-convolution layer by (a) reducing the number of channels in de-convolution layers to avoid the noisy and repeated channels; 
and (b) then creating the reduced channels by learning from their neighboring channels which were previously created in the same block to increase the details in reconstructed face images. 
To train the \textit{NbNets}, a large number of face images are required.
Instead of following the time-consuming and expensive process of collecting a sufficiently large face dataset~\cite{wang2016facescale,otto2016clustering}, we trained a face image generator, DCGAN~\cite{radford2015unsupervised}, to augment available public domain face datasets. 
To further enhance the quality of reconstructed images, we explore both pixel difference and perceptual loss~\cite{johnson2016perceptual} for training the \textit{NbNets}.
In summary, this paper makes following contributions:

$\bullet$ An investigation of the invertibility of face templates generated by deep networks. 
To the best of our knowledge, this is the first such study on security and privacy of face recognition systems.

$\bullet$ An \textit{NbNet} with its building block, NbBlock, was developed for reconstructing face images from deep templates. 
The \textit{NbNets} were trained by data augmentation and perceptual loss~\cite{johnson2016perceptual}, resulting in maintaining discriminative information in deep templates.

$\bullet$ Empirical results show that the proposed face image reconstruction from the corresponding templates is successful. 
We show that we can achieve the following, 
verification rates (\textit{security}), TAR of 95.20\% (58.05\%) on LFW under type-I (type-II) attack @ FAR of 0.1\%.  
For identification (\textit{privacy}), we achieve 96.58\% and 92.84\% rank one accuracy (partition \textit{fa}) in color FERET \cite{phillips2000feret}  as gallery and the images reconstructed from partition \textit{fa} (type-I attack) and \textit{fb} (type-II attack) as probe.

\section{Related Work}
\label{sec:related}

In this section, we describe the current practice of storing face templates, the limitations of current methods for reconstructing face images from deep templates and introduce GANs for generating (synthesizing)  face images.

\subsection{Face Template Security}
Unlike traditional passwords that can be matched in their encrypted or hash form with standard ciphers (e.g., AES, RSA, and SHA-3), face templates cannot be simply protected by standard ciphers because of the intra-subject variations in face images \cite{nandakumar2015biometric,jain201650}. 
Due to the avalanche effect\footnote{https://en.wikipedia.org/wiki/Avalanche\_effect} \cite{stallings2016cryptography} of standard ciphers, the face templates protected by standard ciphers need to be decrypted before matching. 
This introduces another challenge, (decryption) key management. 
In addition, decrypted face templates can also be gleaned by launching an authentication attempt. 

Face template protection remains an open challenge. 
To our knowledge, either the vendors ignore the security and privacy issues of face templates, or secure the encrypted templates and the corresponding keys in specific hardware (e.g., Secure Enclave on A11 of iPhone X \cite{apple2017face}, TrustZone on ARM\footnote{\url{https://www.arm.com/products/security-on-arm/trustzone}}). 
Note that the requirement of specific hardware limits the range of biometric applications. 

\subsection{Reconstructing Face Images from Deep Templates}
\label{sec:relatedrec}

Face template reconstruction requires the determination of the inverse of deep models used to extract deep templates from face images. 
Most deep models are complex and are typically implemented by designing and training a network with sufficiently large capacity~\cite{Goodfellow-et-al-2016}.

\textbf{Shallow model based}~\cite{mohanty2007scores,mignon2013reconstructing}: There are two shallow model based methods for reconstructing face images from templates proposed in the literature: multidimensional scaling (MDS)~\cite{mohanty2007scores} and radial basis function (RBF) regression~\cite{mignon2013reconstructing}. 
However, these methods have only been evaluated using shallow templates. 
The MDS-based method~\cite{mohanty2007scores} uses a set of face images to generate a similarity score matrix using the target face recognition system and then finds an affine space in which face images can approximate the original similarity matrix.
Once the affine space has been found, a set of similarities is obtained from the target face recognition system by matching the target template and the test face images. 
The affine representation of the target template is estimated using these similarities, which is then mapped back to the target face image.

\textbf{Deep model based}\cite{zhmoginov2016inverting,cole2017synthesizing}: 
Zhmoginov and Sandler \cite{zhmoginov2016inverting} learn the reconstruction of face images from templates using a CNN by minimizing the template difference between original and reconstructed images. 
This requires the gradient information from target template extractor and cannot satisfy our assumption of \textit{black-box} template extractor. 
Cole et. al.~\cite{cole2017synthesizing} first estimate the landmarks and textures of face images from the templates, and then combine the estimated landmarks and textures using the differentiable warping to yield the reconstructed images. 
High-quality face images (e.g., front-facing, neutral-pose) are required to be selected for generating landmarks and textures in \cite{cole2017synthesizing} for training the reconstruction model.
Note that both \cite{zhmoginov2016inverting} and \cite{cole2017synthesizing} does not aim to study vulnerability on deep templates and hence no comparable statistical results based template reconstruction attack were reported.

\begin{figure*}[t]
  \centering
  \includegraphics[width=.84\linewidth]{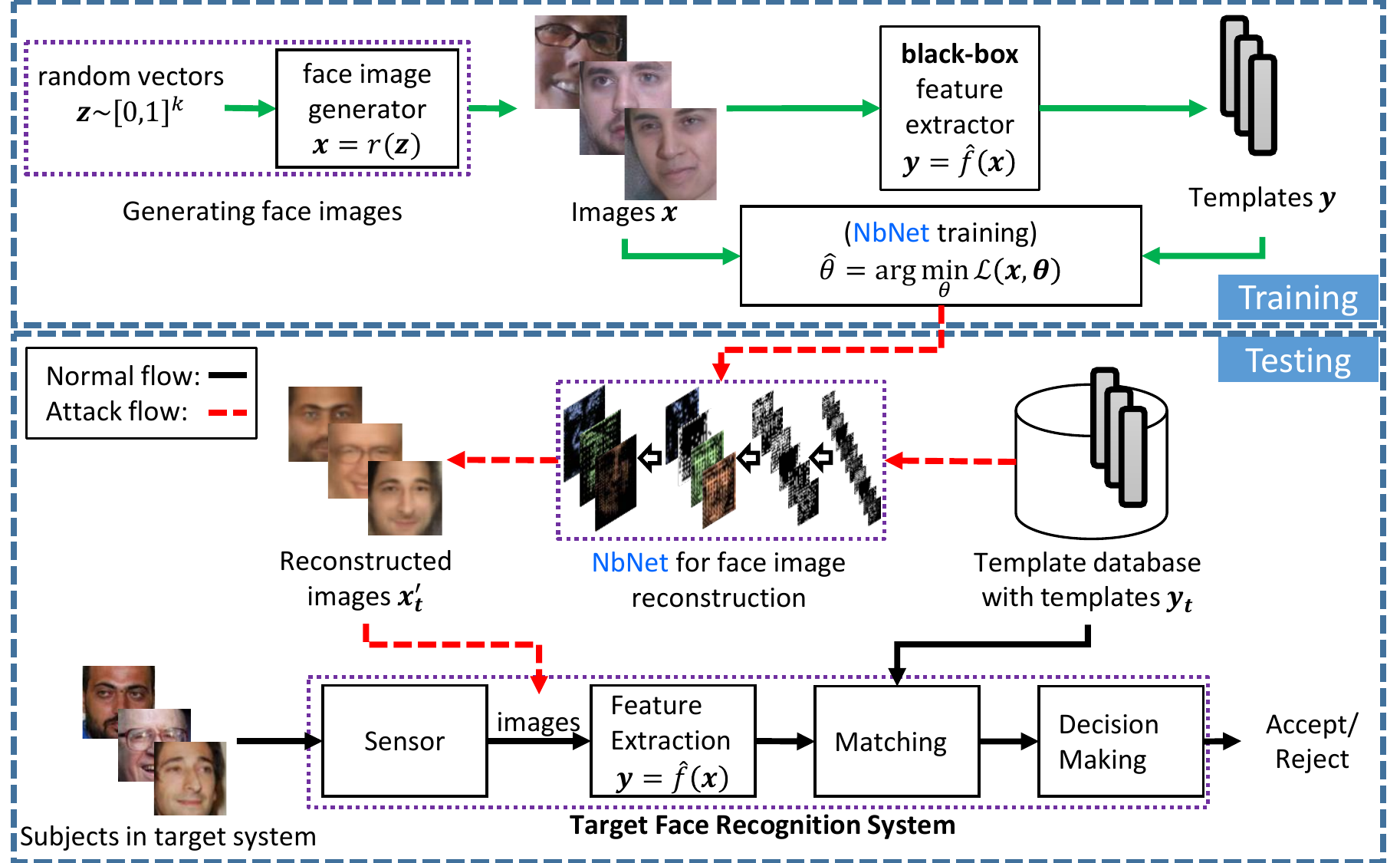}
  \caption{An overview of the proposed system for reconstructing face images from the corresponding deep templates.}
  \label{fig:mthdoverview}
\end{figure*}

\subsection{GAN for Face Image Generation}

With adversarial training, GANs \cite{goodfellow2014generative,goodfellow2016nips,radford2015unsupervised,salimans2016improved,nguyen2016plug,arjovsky2017wasserstein,gulrajani2017improved,berthelot2017began,Mao_2017_ICCV,karras2017progressive} are able to generate photo-realistic (face) images from randomly sampled vectors. 
It has become one of the most popular methods for generating face images, compared to other methods such as data augmentation \cite{masi2016we} and SREFI \cite{banerjee2017srefi}. 
GANs typically consist of a generator which produces an image from a randomly sampled vector, and a discriminator which classifies an input image as real or synthesized. 
The basic idea for training a GAN is to prevent images output by the generator be mistakenly classified as real by co-training a discriminator.

DCGAN~\cite{radford2015unsupervised} is believed to be the first method that directly generates high-quality images ($64\times 64$) from randomly sampled vectors. 
PPGN~\cite{nguyen2016plug} was proposed to conditionally generate high-resolution images with better image quality and sample diversity,  
but it is rather complicated.  
Wasserstein GAN~\cite{arjovsky2017wasserstein,gulrajani2017improved}  was proposed to solve the model collapse problems in GAN \cite{goodfellow2016nips}.
Note that the images generated by Wasserstein GAN~\cite{arjovsky2017wasserstein,gulrajani2017improved} are comparable with those output by DCGAN. 
BEGAN~\cite{berthelot2017began} and LSGAN \cite{Mao_2017_ICCV} have been proposed to attempt to address the model collapse, and non-convergence problems with GAN.
A progressive strategy for training high-resolution GAN is described in \cite{karras2017progressive}.

In this work, we employed an efficient yet effective method, DCGAN to generate face images. 
The original DCGAN~\cite{radford2015unsupervised} is easy to collapse and outputs poor quality high-resolution images (e.g., $160\times 160$ in this work). 
We address the above problems with DCGAN (Section \ref{sec:training_details}).

\section{Proposed Template Security Study}
\label{sec:proposed}

An overview of our security study for deep template based face recognition systems under template reconstruction attack is shown in Fig.~\ref{fig:mthdoverview}; the normal processing flow and template reconstruction attack flows are shown as black solid and red dotted lines, respectively.
This section first describes the scenario of template reconstruction attack using an adversarial machine learning framework~\cite{biggio2015adversarial}.
This is followed by the proposed \textit{NbNet} for reconstructing face images from deep templates and the corresponding training strategy and implementation. 

\begin{figure*}[t]
  \begin{minipage}{0.58\linewidth}
    \centering
    \includegraphics[width=.8\linewidth]{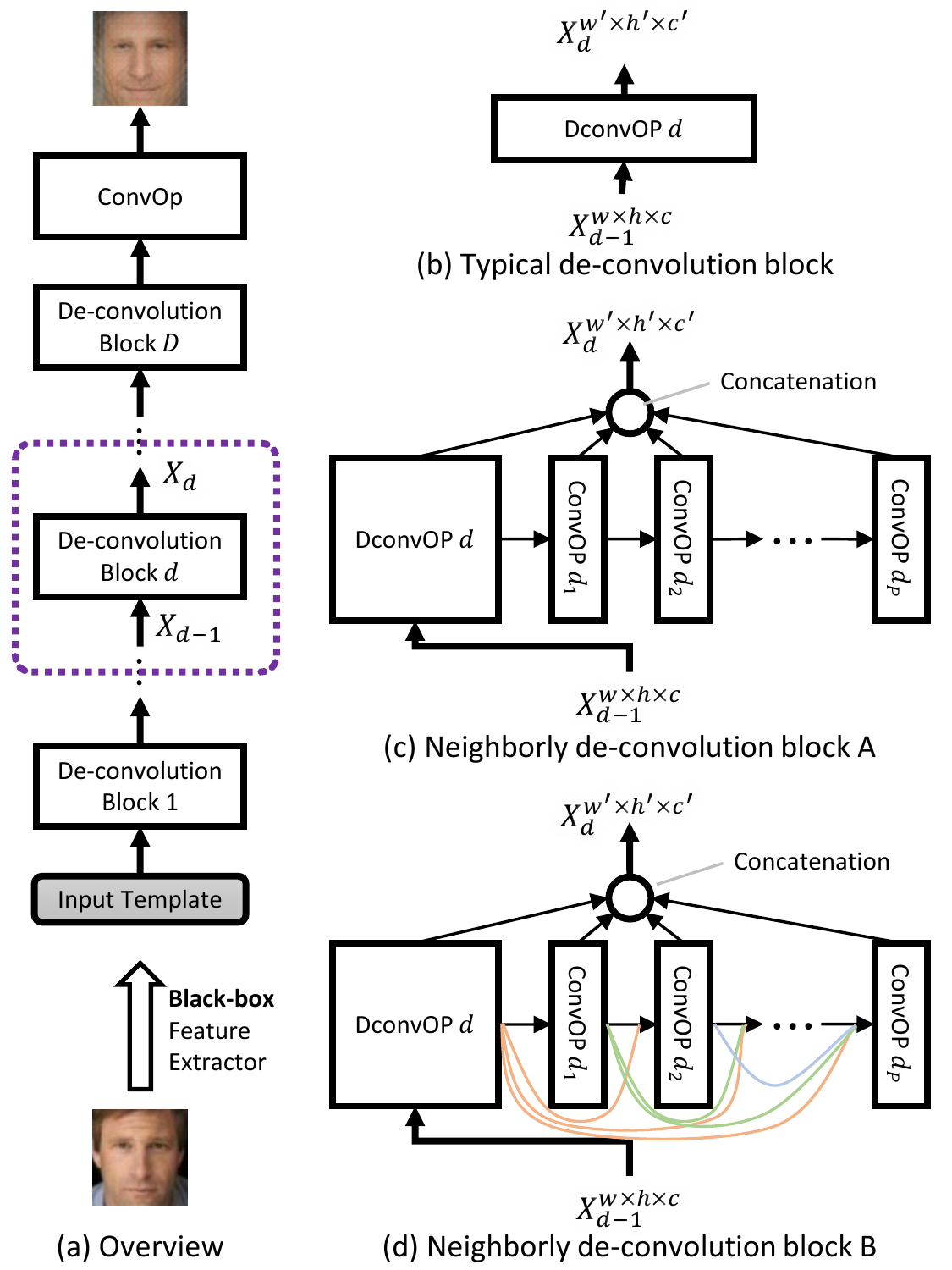}
    \caption{The proposed \textit{NbNet} for reconstructing face images from the corresponding face templates. (a) Overview of our face reconstruction network, (b) typical de-convolution block for building de-convolutional neural network (D-CNN), (c) and (d) are the neighborly de-convolution blocks (NbBlock) A/B for building NbNet-A and NbNet-B, respectively.
    Note that ConvOP (DconvOP) denotes a cascade of a convolution (de-convolution), a batch-normalization~\cite{ioffe2015batch}, and a ReLU activation (tanh in ConvOP of (a)) layers, where the width of ConvOp (DconvOP) denotes the number of channels in its  convolution (de-convolution) layer. 
    The black circles in (c) and (d) denote the channel concatenation of the output channels of DconvOP and ConvOPs. 
    }
    \label{fig:mthddcnn}
  \end{minipage}
  \hfill
  \begin{minipage}{0.4\linewidth}
    \centering
      \subfloat[D-CNN]{{\includegraphics[width=.8\linewidth]{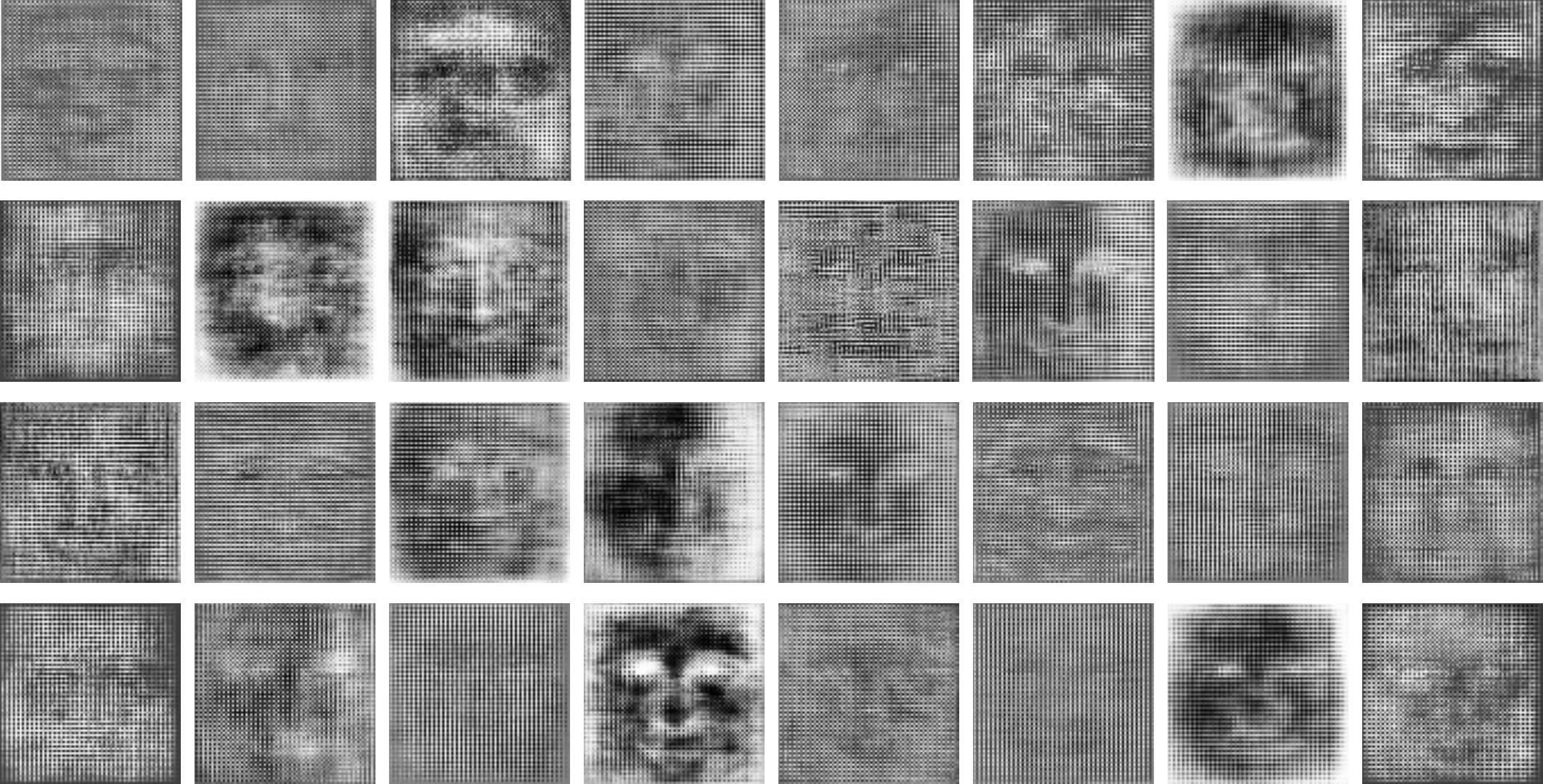}} 
          \label{fig:layer4_dconvnn}}\\
      \subfloat[NbNet-A]{{\includegraphics[width=.8\linewidth]{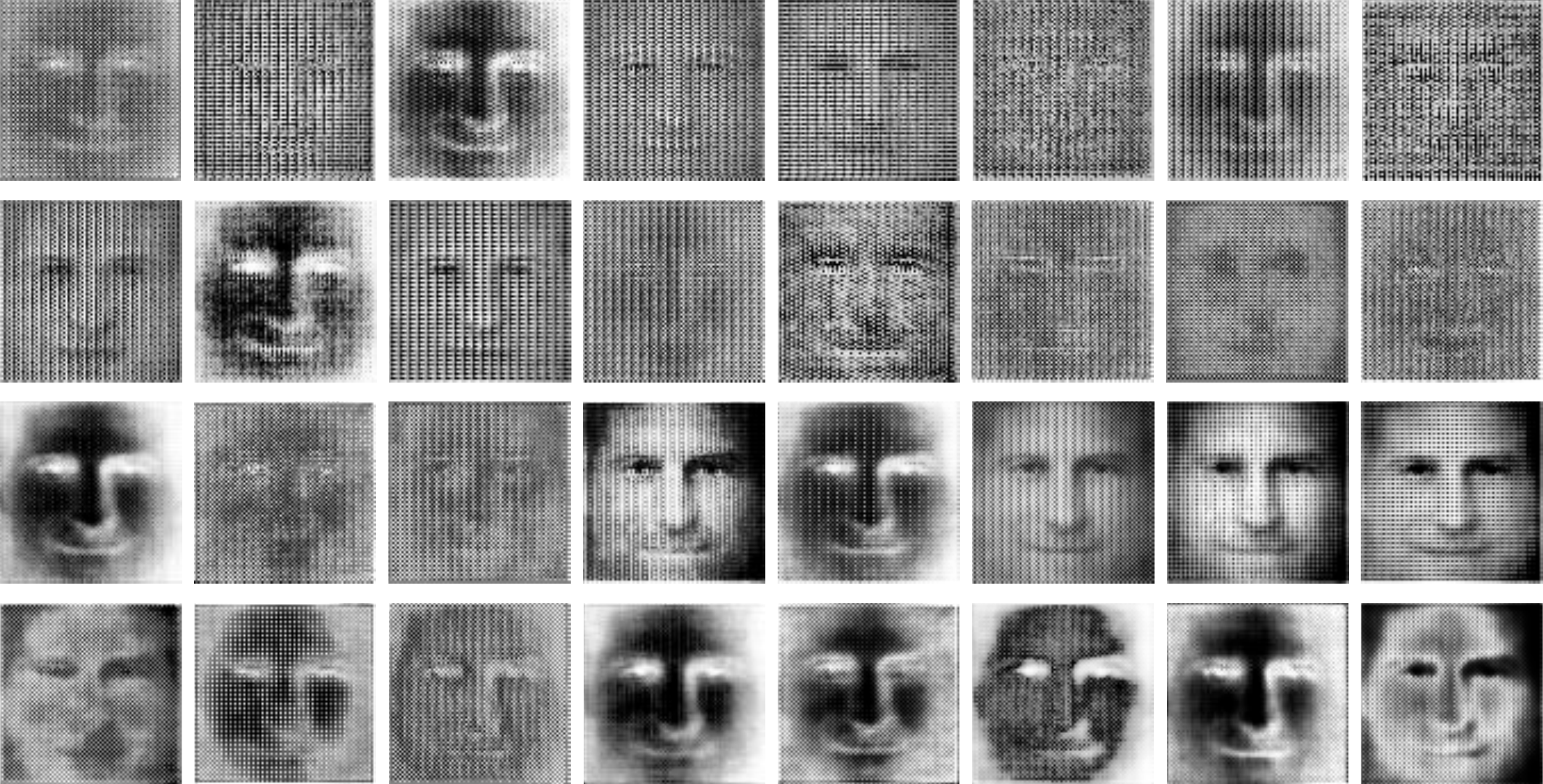}} 
       \label{fig:layer4_nbneta}}\\
      \subfloat[NbNet-B]{{\includegraphics[width=.8\linewidth]{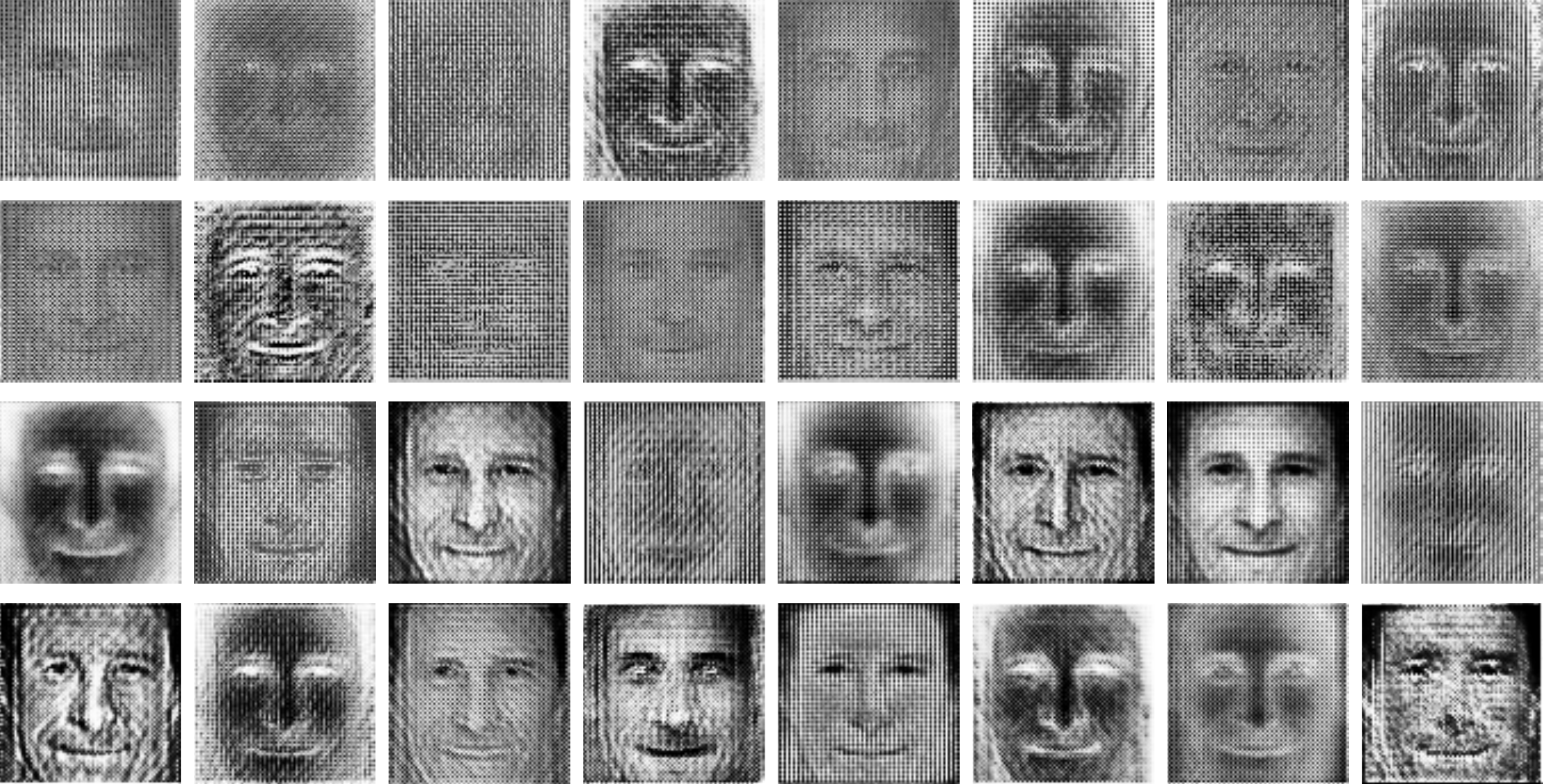}} 
       \label{fig:layer4_nbnetb}}\\
      \caption{Visualization of 32 output channels of the \textit{5th} de-convolution blocks in (a) D-CNN, (b) NbNet-A, and (c) NbNet-B networks, where the input template was extracted from the bottom image of Fig.~\ref{fig:mthddcnn}~(a). 
      Note that the four rows of channels in (a) and the first two rows of channels in (b) and (c) are learned from channels from the corresponding \textit{4th} block. 
      The third row of channels in both (b) and (c) are learned from their first two rows of channels. 
      The fourth row of channels in (b) is learned from the third row of channels only, where the fourth row of channels in (c) is learned from the first three rows of channels.}
      \label{fig:layer4}
  \end{minipage}
\end{figure*}

\subsection{Template Reconstruction Attack}

The adversarial machine learning framework~\cite{biggio2015adversarial,mai2017guessability} categorizes biometric attack scenarios from four perspectives: an adversary's goal and his/her  knowledge, capability, and attack strategy. 
Given a deep template based face recognition system, 
our template reconstruction attack scenario using the adversarial machine learning framework is as follows.

  $\bullet$ \textit{Adversary's goal:} The attacker aims to impersonate a subject in the target face recognition system, compromising the system integrity.
  
  $\bullet$ \textit{Adversary's knowledge:} The attacker is assumed to have the following information. (a) The templates $\bm{y}_t$ of the target subjects, which can be obtained via template database leakage or an insider attack. 
  (b) The \textit{black-box} feature extractor $\bm{y} = \hat{f}(\bm{x})$ of the target face recognition system. This can potentially be obtained by purchasing the target face recognition system's SDK. 
  The attacker has neither information about target subjects nor their enrollment environments. Therefore, no face images enrolled in the target system can be utilized in the attack.
  
  $\bullet$ \textit{Adversary's capability:}(a) Ideally, the attacker should only be permitted to present fake faces (2D photographs or 3D face masks) to the face sensor during authentication.
  In this study, to simplify, the attacker is assumed to be able to inject face images directly into the feature extractor as if the images were captured by the face sensor. 
  Note that the injected images could be used to create fake faces in actual attacks. 
  (b) The identity decision for each query is available to the attacker. However, the similarity score of each query cannot be accessed. 
  (c) Only a small number of trials (e.g., $<5$) are permitted for the recognition of a target subject.
  
  $\bullet$ \textit{Attack strategy:} Under these assumptions, the attacker can infer a face image $\bm{x}_t$ from the target template $\bm{y}_t$ using a reconstruction model $\bm{x}_t = g_{\bm\theta} (\bm{y}_t)$ and insert reconstructed image as a query to access the target face recognition system. 
  The parameter $\bm{\theta}$ of the reconstruction model $g_{\bm{\theta}}(\cdot)$ can be learned using public domain face images. 

\subsection{\textit{NbNet} for Face Image Reconstruction}
\label{sec:proposed_dcnn}

\subsubsection{Overview}
\label{sec:propose_overview}

An overview of the proposed \textit{NbNet} is shown in Fig.~\ref{fig:mthddcnn}~(a). 
The \textit{NbNet} is a cascade of multiple stacked de-convolution blocks and a convolution operator, ConvOp. 
De-convolution blocks up-sample and expand the abstracted signals in the input channels to produce output channels with a larger size as well as more details about reconstructed images. 
With multiple ($D$) stacked de-convolution blocks, the \textit{NbNet} is able to expand highly abstracted deep templates back to channels with high resolutions and sufficient details for generating the output face images. 
The ConvOp in Fig.~\ref{fig:mthddcnn}~(a) aims to summarize multiple output channels of $D$-th de-convolution block to the target number of channels (3 in this work for RGB images).
It is a cascade of convolution, batch-normalization~\cite{ioffe2015batch}, and tanh activation layers. 

\subsubsection{Neighborly De-convolution Block}
\label{sec:propose_nbblock}

A typical design of the de-convolution block~\cite{zeiler2010deconvolutional,radford2015unsupervised}, as shown in Fig.~\ref{fig:mthddcnn}~(b), is to learn output channels with up-sampling from channels in previous blocks only. 
The number of output channels $c'$ is often made large enough to ensure sufficient model capacity for template reconstruction \cite{zagoruyko2016wide}. 
However, the up-sampled output channels tend to suffer from the following two issues:
(a) noisy and repeated channels; and 
(b) insufficient details. 
An example of these two issues is shown in Fig.~\ref{fig:layer4}~(a), which is a visualization of output channels in the fifth de-convolution block of a D-CNN that is built with typical de-convolution blocks.
The corresponding input template was extracted from the bottom image of Fig.~\ref{fig:mthddcnn}~(a).

To address these limitations, we propose NbBlock which produces the same number of output channels as typical de-convolution blocks for the face template reconstruction.
One of the reasons for noisy and repeated output channels is that a large number of channels are treated equally in a typical de-convolution block; from the perspective of network architecture, these output channels were learned from the same set of input channels and became the input of the same forthcoming blocks. 
To mitigate this issue, we first reduce the number of output channels that is simultaneously learned from the previous blocks. 
We then create the reduced number of output channels with enhanced details by learning from neighbor channels in the same block.

Let $G_d(\cdot)$ denote the $d$-th NbBlock, which is shown as the dashed line in Fig.~\ref{fig:mthddcnn}~(a) and is the building component of our \textit{NbNet}. 
Suppose that $G_d(\cdot)$ consists of one de-convolution operator (DconvOP) $\mathcal{N}'_d$ and $P$ convolution operators (ConvOPs) $\{\mathcal{N}_{d_p}| p=1,2,\cdots, P\}$. 
Let $X'_d$ and $X_{d,p}$ denote the output of DconvOP $\mathcal{N}'_d$ and $p$-th ConvOP $\mathcal{N}_{d_p}$ in $d$-th NbBlock $G_d(\cdot)$, then 
\begin{equation}
X_d = G_d(X_{d-1}) = [\mathcal{X}_P]
\label{eq:devblock}
\end{equation}
where $X_{d-1}$ denotes the output of the $(d-1)$-th NbBlock, $[\cdot]$ denotes a function of channel concatenation, and $\mathcal{X}_P$ is the set of outputs of DconvOP and all ConvOPs in $G_d(\cdot)$, 
\begin{equation}
\mathcal{X}_P=\{X'_d, X_{d,1}, X_{d,2}, \cdots, X_{d,P}\}
\end{equation}
where $X'_d$ and $X_{d,p}$ denotes the output of DconvOP and the $p$-th ConvOP in $d$-th block, resp., and satisfy
\begin{equation}
X'_d=\mathcal{N'}_d\left(X_{d-1}\right), X_{d,p}=\mathcal{N}_{d_p}\left([\hat{\mathcal{X}_p}]\right)
\end{equation}
where $\hat{\mathcal{X}_p}$ is a non-empty subset of ${\mathcal{X}_p}$.

Based on this idea, we built two NbBlocks, A and B, as shown in Figs.~\ref{fig:mthddcnn}~(c) and~(d), where the corresponding reconstructed networks are named NbNet-A and NbNet-B, respectively. 
In this study, the DconvOp (ConvOp) in Figs.~\ref{fig:mthddcnn}~(b), (c), and (d) denotes a cascade of de-convolution (convolution), batch-normalization~\cite{ioffe2015batch}, and ReLU activation layers. 
The only difference between blocks A and B is the choice of $\hat{\mathcal{X}_p}$,
\begin{equation}
    \begin{split}
    \hat{\mathcal{X}_p} = 
    &\left\{
    \begin{aligned}
    &\{X_{d}'\},&\text{blocks A \& B with } p=1;\\
    &\{X_{d,p-1}\},&\text{block A with } p>1;\\
    &{\mathcal{X}_p},&\text{block B with } p>1.\\
    \end{aligned}
    \right.
    \end{split}
\end{equation}
In our current design of the NbBlocks, half of output channels ($\frac{c'}{2}$ for block $d$) are produced by a DconvOP, and the remaining channels are produced by $P$ ConvOPs, each of which gives, in this study, eight output channels (Table.~\ref{tb:netarchi}). 
Example of blocks A and B with 32 output channels are shown in Figs.~\ref{fig:layer4}~(b) and (c). 
The first two rows of channels are produced by DconvOp and the third and fourth rows of channels are produced by the first and second ConvOps, respectively. 
Compared with Fig.~\ref{fig:layer4}~(a), the first two rows in Figs.~\ref{fig:layer4}~(b) and~(c) have small amount of noise and fewer repeated channels, where the third and fourth row provide channels with more details about the target face image (the reconstructed image in Fig.~\ref{fig:mthddcnn}~(a)).
The design of our NbBlocks is motivated by \textit{DenseNet}  \cite{huang2016densely} and \textit{MemNet} \cite{tai2017memnet}. 

\subsection{Reconstruction Loss}
\label{sec:proposed_loss}

Let us denote $\mathcal{R}\left(\bm{x},\bm{x}'\right)$ as the reconstruction loss between an input face image $\bm x$ and its reconstruction ${\bm x}' = g_{\bm\theta}\left(\hat{f}(\bm{x})\right)$, where $g_{\bm\theta}(\cdot)$ denotes an \textit{NbNet} with parameter $\bm \theta$ and $\hat{f}(\cdot)$ denotes a \textit{black-box} deep template extractor.

\textbf{Pixel Difference}:
A straightforward loss for learning reconstructed image ${\bm x}'$ is pixel-wise loss between ${\bm x}'$ and its original version ${\bm x}$. The Minkowski distance could then be used and mathematically expressed as  
\begin{equation}
\mathcal{R}_{pixel}\left(\bm{x},\bm{x}'\right) = \left|\left|\bm{x} - \bm{x}' \right|\right|_k 
= \left(\sum_{m=1}^{M} \left|x_m - x_m'\right|^k\right)^{\frac{1}{k}}
\end{equation}
where $M$ denotes number of pixels in $\bm x$ and $k$ denotes the order of the metric.

\textbf{Perceptual Loss} \cite{johnson2016perceptual}:
Because of the high discriminability of deep templates, most of the intra-subject variations in a face image might have been eliminated in its corresponding deep template.
The pixel difference based reconstruction leads to a difficult task of reconstructing these eliminated intra-subject variations, which, however, are not necessary for reconstruction. 
Besides, it does not consider holistic contents in an image as interpreted by machines and human visual perception. 
Therefore, instead of using pixel difference, we employ the perceptual loss \cite{johnson2016perceptual} which guides the reconstructed images towards the same representation as the original images. 
Note that a good representation is robust to intra-subject variations in the input images. 
The representation used in this study is the feature map in the VGG-19 model \cite{simonyan2014very}\footnote{Provided by \url{https://github.com/dmlc/mxnet-model-gallery}}. 
We empirically determine that using the output of \textit{ReLU3\_2} activation layer as the feature map leads the best image reconstruction, in terms of face matching accuracy. 
Let $F(\cdot)$ denote feature mapping function of the \textit{ReLU3\_2} activation layer of VGG-19 \cite{simonyan2014very}, then the perceptual loss can be expressed as
\begin{equation}
\mathcal{R}_{percept}\left(\bm{x},\bm{x}'\right) = \frac{1}{2}\left|\left|F(\bm{x}) - F(\bm{x}') \right|\right|_2^2
\end{equation}

\begin{table*}[!t]
    \caption{Network details for D-CNN and \textit{NbNets}. ``$[k_1\times k_2, c]$ DconvOP (ConvOP), stride $s$'', denotes cascade of a de-convolution (convolution) layer with $c$ channels, kernel size ($k_1,k_2$) and stride $s$, batch normalization, and ReLU (tanh for the bottom ConvOP) activation layer.}
    \centering
    \setlength{\extrarowheight}{.4mm}
    \begin{tabular*}{.8\linewidth}{M{0.13\linewidth}|M{0.1\linewidth}|M{0.23\linewidth}|M{0.25\linewidth}}
    \hline
    Layer name & Output size ($c\times w\times h$) & D-CNN & NbNet-A, NbNet-B \\ \hline
    input layer & $128\times 1 \times 1$\\\hline
    De-convolution Block (1) & $512\times 5\times 5$ &$[5\times 5, 512]$ DconvOP, stride 2 & $[5\times 5, 256]$ DconvOP, stride 2  \{$[3\times 3, 8]$ ConvOP, stride 1\}$\times 32$\\\hline
    De-convolution Block (2) & $256\times 10\times 10$ &$[3\times 3, 256]$ DconvOP, stride 2 & $[3\times 3, 128]$ DconvOP, stride 2  \{$[3\times 3, 8]$ ConvOP, stride 1\}$\times 16$\\\hline
    De-convolution Block (3) & $128\times 20\times 20$ &$[3\times 3, 128]$ DconvOP, stride 2 & $[3\times 3, 64]$ DconvOP, stride 2  \{$[3\times 3, 8]$ ConvOP, stride 1\}$\times 8$\\\hline
    De-convolution Block (4) & $64\times 40\times 40$ &$[3\times 3, 64]$ DconvOP, stride 2 & $[3\times 3, 32]$ DconvOP, stride 2  \{$[3\times 3, 8]$ ConvOP, stride 1\}$\times 4$\\\hline
    De-convolution Block (5) & $32\times 80\times 80$ &$[3\times 3, 32]$ DconvOP, stride 2 & $[3\times 3, 16]$ DconvOP, stride 2  \{$[3\times 3, 8]$ ConvOP, stride 1\}$\times 2$\\\hline
    De-convolution Block (6) & $16\times 160\times 160$ &$[3\times 3, 16]$ DconvOP, stride 2 & $[3\times 3, 8]$ DconvOP, stride 2  \{$[3\times 3, 8]$ ConvOP, stride 1\}$\times 1  $\\\hline
    ConvOP & $3\times 160\times 160$ &\multicolumn{2}{c}{$[3\times 3,3]$ ConvOP, stride 1}\\\hline
    Loss layer & $3\times 160\times 160$ &\multicolumn{2}{c}{Pixel difference or perceptual loss \cite{johnson2016perceptual}}\\\hline
    \end{tabular*}
    \label{tb:netarchi}
\end{table*}

\subsection{Generating Face Images for Training}
\label{sec:proposed_strategy}

To successfully launch an template reconstruction attack on a face recognition system without knowledge of the target subject population, \textit{NbNets} should be able to accurately reconstruct face images with input templates extracted from face images of different subjects.
Let $p_{\bm{x}}(\bm{x})$ denote the probability density function (pdf) of image $\bm{x}$, the objective function for training a \textit{NbNet} can be formulated as 
\begin{equation}
\begin{split}
  \arg \min_{\bm\theta} \mathcal{L}\left(\bm{x},\bm{\theta}\right) 
  &= \arg \min_{\bm\theta} \int \mathcal{R}\left(\bm{x},\bm{x}'\right) p_{\bm{x}}(\bm{x}) d\bm{x}\\
  = \arg \min_{\bm\theta} &\int \mathcal{R}\left(\bm{x},g_{\bm\theta}(\hat{f}(\bm{x}))\right) p_{\bm{x}}(\bm{x}) d\bm{x}.
\end{split}
\label{eq:idealobj}
\end{equation}

Since there are no explicit methods for estimating $p_{\bm{x}}(\bm{x})$, we cannot sample face images from $p_{\bm{x}}(\bm{x})$.
The common approach is to collect a large-scale face dataset and approximate the loss function $\mathcal{L}({\bm\theta})$ in Eq.~(\ref{eq:idealobj}) as:
\begin{equation}
  \mathcal{L}(\bm{x},{\bm\theta}) = 
  \frac{1}{N}\sum_{i}^{N} \mathcal{R}\left(\bm{x}_i,g_{\bm\theta}(\hat{f}(\bm{x}_i))\right)
  \label{eq:approobj}
\end{equation}
where $N$ denotes the number of face images and $\bm{x}_i$ denotes the $i$-th training image.
This approximation is optimal if, and only if, $N$ is sufficiently large. 
In practice, this is not feasible because of the huge time and cost associated with collecting a large database of face images.

To train a generalizable \textit{NbNet} for reconstructing face images from their deep templates, a large number of face images are required.
Ideally, these face images should come from a large number of different subjects because deep face templates of the same subject are very similar and can be regarded as either single exemplar or under large intra-user variations, a small set of exemplars in the training of \textit{NbNet}. 
However, current large-scale face datasets (such as VGG-Face~\cite{Parkhi15}, CASIA-Webface~\cite{yi2014learning}, and Multi-PIE~\cite{gross2007cmu}) were primarily collected for training or evaluating face recognition algorithms.
Hence, they either contain an insufficient number of images (for example, 494K images in CASIA-Webface) or an insufficient number of subjects (for instance, 2,622 subjects in VGG-Face and 337 subjects in Multi-PIE) for training a reconstruction \textit{NbNet}.

Instead of collecting a large face image dataset for training, we propose to augment current publicly available datasets.
A straightforward way to augment a face dataset is to estimate the distribution of face images $p_{\bm{x}}(\bm{x})$ and then sample the estimated distribution. 
However, as face images generally consist of a very large number of pixels, there is no efficient method to model the joint distribution of these pixels. 
Therefore, we introduced a generator $\bm{x}=r(\bm{z})$ capable of generating a face image $\bm{x}$ from a vector $\bm{z}$ with a given distribution.
Assuming that $r(\bm{z})$ is one-to-one and smooth, the face images can be sampled by sampling $\bm{z}$.
The loss function $\mathcal{L}(\bm{\theta})$ in Eq.~(\ref{eq:idealobj}) can then be approximated as follows:
\begin{equation}
\begin{split}
\mathcal{L}\left(\bm{x},\bm{\theta}\right)  &= 
\int \mathcal{R}\left(\bm{x},g_{\bm\theta}(\hat{f}(\bm{x}))\right) p_{\bm{x}}(\bm{x}) d\bm{x} \\
&= \int \mathcal{R}\left(r(\bm{z}),g_{\bm\theta}\left(\hat{f}(r(\bm{z}))\right)\right) p_{\bm{z}}(\bm{z}) d\bm{z}. \\
\end{split}
\label{eq:dcganappobj}
\end{equation}
where $p_{\bm{z}}(\bm{z})$ denotes the pdf of variable $\bm{z}$. Using the \textit{change of variables} method~\cite{random2017,dokuchaev2015probability},  it is easy to show that $p_{\bm{z}}(\bm{z})$ and $r(\bm{z})$ have the following connection,
\begin{equation}
\begin{split}
p_{\bm{z}}(\bm{z}) = p_{\bm{x}}(r(\bm{z}))\left|\text{det}\left(\frac{d\bm{x}}{d\bm{z}}\right)\right|, \text{where} \left(\frac{d\bm{x}}{d\bm{z}}\right)_{ij} = \frac{\partial x_i}{\partial z_j}
\end{split}.
\end{equation}

Suppose a face image $\bm{x}\in \mathbb{R}^{h\times w \times c}$ of height $h$, width $w$, and with $c$ channels can be represented by a real vector $\bm{b}=\{b_1,\cdots,b_k\}\in \mathbb{R}^k$ in a manifold space with $h\times w\times c \gg k$.
It can then be shown that there exists a generator function $\bm{b'}=\hat{r}(\bm{z})$ that generates $\bm{b'}$ with a distribution identical to that of $\bm{b}$,
where $\bm{b}$ can be arbitrarily distributed and $\bm{z} \in [0,1]^k$ is uniformly distributed (see Appendix).

To train the \textit{NbNets} in the present study, we used the generative model of a DCGAN~\cite{radford2015unsupervised} as our face generator $r(\cdot)$. 
This model can generate face images from vectors $\bm{z}$ that follow a uniform distribution.
Specifically, DCGAN generates face images $r(\bm{z})$ with a distribution that is an approximation to that of real face images $\bm{x}$.  
It can be shown empirically that a DCGAN can generate face images of unseen subjects with different intra-subject variations.
By using adversarial learning, the DCGAN is able to generate face images that are classified as real face images by a co-trained real/fake face image discriminator.
Besides, the intra-subject variations generated using a DCGAN can be controlled by performing arithmetic operations in the random input space~\cite{radford2015unsupervised}.

\subsection{Differences with \textit{DenseNet} }
One of the related work to \textit{NbNet} is \textit{DenseNet} \cite{huang2016densely}, from which the \textit{NbNet} is inspired. 
Generally, \textit{DenseNet} is based on convolution layers and designed for object recognition, while the proposed \textit{NbNet} is based on de-convolution layers and aimed to reconstruct face images from deep templates. 
Besides, \textit{NbNet} is a framework whose NbBlocks produce output channels learned from previous blocks and neighbor channels within the block. 
The output channels of NbBlocks consist of fewer repeated and noisy channels and contain more details for face image reconstruction than the typical de-convolution blocks. 
Under the framework of \textit{NbNet}, one could build a skip-connection-like network \cite{he2016deep}, NbNet-A, and a \textit{DenseNet} -like network, NbNet-B.
Note that NbNet-A sometimes achieves a comparable performance to NbNet-B with roughly 67\% of the parameters and 54\% running time only (see model VGG-NbA-P and VGG-NbB-P in Section~\ref{sec:experiment}). 
We leave more efficient and accurate \textit{NbNets} construction as a future work.

\subsection{Implementation Details}
\label{sec:proposed_train}
\subsubsection{Network Architecture}
\noindent The detailed architecture of the D-CNN and the proposed \textit{NbNets} is shown in Table.~\ref{tb:netarchi}. 
The NbNet-A and NbNet-B show the same structure in Table.~\ref{tb:netarchi}.
However, the input of the ConvOP in the de-convolution blocks (1)-(6) are different (Fig.~\ref{fig:mthddcnn}), where NbNet-A uses the nearest previous channels in the same block, and NbNet-B uses all the previous channels in the same block.

\subsubsection{Revisiting DCGAN}
\label{sec:training_details}

\begin{figure}[t]
	\centering
	
	\subfloat[VGG-Face]{
		\begin{minipage}{.475\linewidth}
			{\includegraphics[width=.23\linewidth]{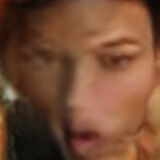}} 
			{\includegraphics[width=.23\linewidth]{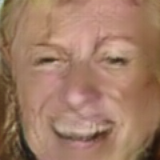}} 
			{\includegraphics[width=.23\linewidth]{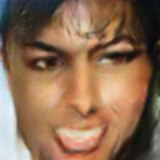}}
			{\includegraphics[width=.23\linewidth]{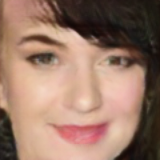}} \vspace{0.1cm}\\
			{\includegraphics[width=.23\linewidth]{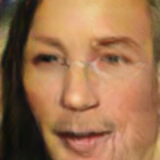}}
			{\includegraphics[width=.23\linewidth]{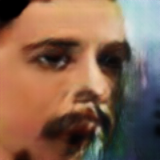}}
			{\includegraphics[width=.23\linewidth]{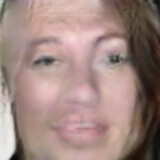}}
			{\includegraphics[width=.23\linewidth]{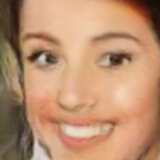}}
		\end{minipage}
		\label{fig:vgg_genexample}
	}\hfill
	\subfloat[Multi-PIE]{
		\begin{minipage}{.475\linewidth}
			{\includegraphics[width=.23\linewidth]{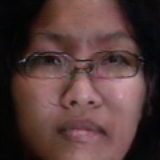}} 
			{\includegraphics[width=.23\linewidth]{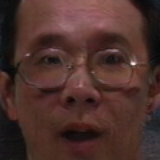}} 
			{\includegraphics[width=.23\linewidth]{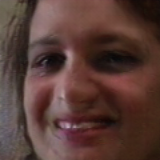}}
			{\includegraphics[width=.23\linewidth]{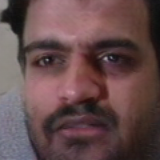}} \vspace{0.1cm}\\
			{\includegraphics[width=.23\linewidth]{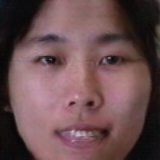}}
			{\includegraphics[width=.23\linewidth]{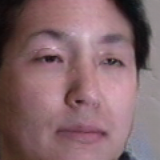}}
			{\includegraphics[width=.23\linewidth]{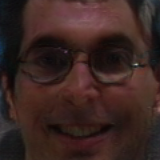}}
			{\includegraphics[width=.23\linewidth]{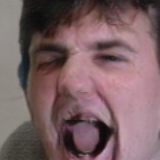}}
		\end{minipage}
		\label{fig:multipie_genexample}
	}  
	\caption{Sample face images generated from face generators trained on (a) VGG-Face, and (b) Multi-PIE.}
	\label{fig:gen_example}
\end{figure}

To train our \textit{NbNet} to reconstruct face images from deep templates, we first train a DCGAN to generate face images. 
These generated images are then used for training.
The face images generated by the original DCGAN could be noisy and sometimes difficult to interpret. 
Besides, the training as described in \cite{radford2015unsupervised} is often collapsed in generating high-resolution images. 
To address these issues, we revisit the DCGAN as below (as partially suggested in \cite{goodfellow2016nips}):
\begin{itemize}
  \item \textit{Network architecture}: replace the batch normalization and ReLU activation layer in both generator and discriminator by the SeLU activation layer \cite{klambauer2017self}, which performs the normalization of each training sample. 
  \item \textit{Training labels}: replace the hard labels (`1' for real, and `0' for generated images) by soft labels in the range [0.7, 1.2] for real, and in range [0, 0.3] for generated images. 
  This helps smooth the discriminator and avoids model collapse. 
  \item \textit{Learning rate}: in the training of DCGAN, at each iteration, the generator is updated with one batch of samples, while the discriminator is updated with two batches of samples (1 batch of `real' and 1 batch of `generated'). 
  This often makes the discriminator always correctly classify the images output by the generator. 
  To balance, we adjust the learning rate of the generator to $2\times 10^{-4}$, which is greater than the learning rate of the discriminator,  $5\times 10^{-5}$.
\end{itemize}
Example generated images were shown in Fig.~\ref{fig:gen_example}.
\begin{figure}[t]
	\centering
	\subfloat[VGG-Face]{
		\begin{minipage}{.475\linewidth}
			{\includegraphics[width=.23\linewidth]{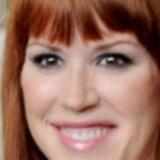}} 
			{\includegraphics[width=.23\linewidth]{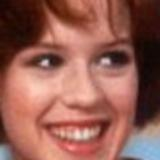}} 
			{\includegraphics[width=.23\linewidth]{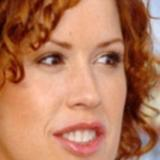}}
			{\includegraphics[width=.23\linewidth]{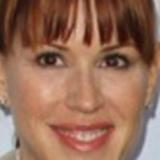}} \vspace{0.1cm}\\
			{\includegraphics[width=.23\linewidth]{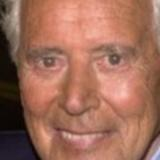}}
			{\includegraphics[width=.23\linewidth]{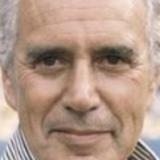}}
			{\includegraphics[width=.23\linewidth]{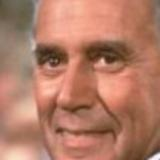}}
			{\includegraphics[width=.23\linewidth]{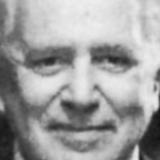}}
		\end{minipage}
		\label{fig:vgg_dbexample}
	}\hfill
	\subfloat[Multi-PIE]{
		\begin{minipage}{.475\linewidth}
			{\includegraphics[width=.23\linewidth]{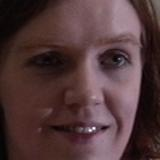}} 
			{\includegraphics[width=.23\linewidth]{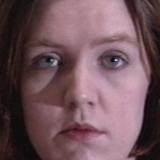}} 
			{\includegraphics[width=.23\linewidth]{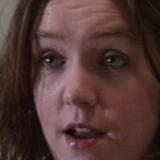}}
			{\includegraphics[width=.23\linewidth]{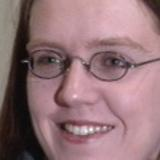}} \vspace{0.1cm}\\
			{\includegraphics[width=.23\linewidth]{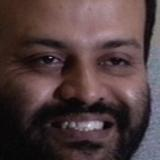}}
			{\includegraphics[width=.23\linewidth]{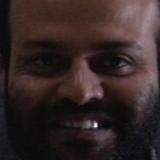}}
			{\includegraphics[width=.23\linewidth]{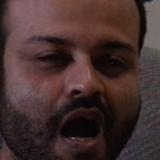}}
			{\includegraphics[width=.23\linewidth]{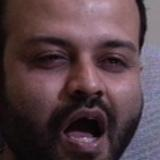}}
		\end{minipage}
		\label{fig:multipie_dbexample}
	} 
	
	\subfloat[LFW]{
		\begin{minipage}{.475\linewidth}
			{\includegraphics[width=.23\linewidth]{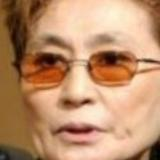}} 
			{\includegraphics[width=.23\linewidth]{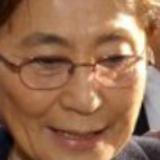}} 
			{\includegraphics[width=.23\linewidth]{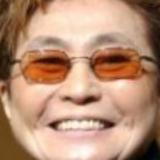}}
			{\includegraphics[width=.23\linewidth]{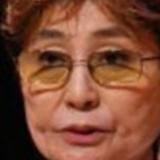}} 
		\end{minipage}
		\label{fig:lfw_dbexample}
	}
	\subfloat[FRGC v2.0]{
		\begin{minipage}{.475\linewidth}
			{\includegraphics[width=.23\linewidth]{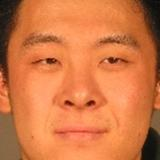}} 
			{\includegraphics[width=.23\linewidth]{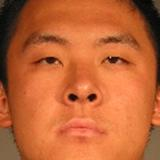}} 
			{\includegraphics[width=.23\linewidth]{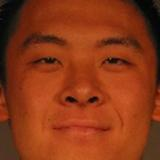}}
			{\includegraphics[width=.23\linewidth]{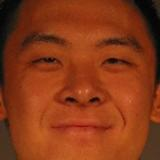}} 
		\end{minipage}
		\label{fig:frgc_dbexample}
	}
	
	\subfloat[Color FERET]{
		\begin{minipage}{.475\linewidth}
			{\includegraphics[width=.23\linewidth]{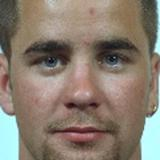}} 
			{\includegraphics[width=.23\linewidth]{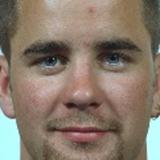}} 
			{\includegraphics[width=.23\linewidth]{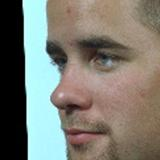}}
			{\includegraphics[width=.23\linewidth]{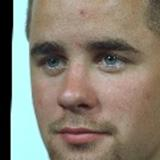}} 
		\end{minipage}
		\label{fig:feret_dbexample}
	}
	\caption{Example face images from the training and testing datasets: (a) VGG-Face (1.94M images)~\cite{Parkhi15}, (b) Multi-PIE (151K images, only three camera views were used, including $`14\_0', `05\_0'$ and $`05\_1'$, respectively)~\cite{gross2007cmu}, (c) LFW (13,233 images)~\cite{Huang2012a,learned2016labeled}, (d) FRGC v2.0 (16,028 images in the target set of Experiment 1)\cite{phillips2005overview}, and (e) Color FERET (2,950 images)~\cite{phillips2000feret}. }
	\label{fig:datasetexample}
\end{figure}

\subsubsection{Training Details}
With the pre-trained DCGAN, face images were first generated by randomly sampling vectors $\bm{z}$ from a uniform distribution and the corresponding face templates were extracted.
The \textit{NbNet} was then updated with the generated face images as well as the corresponding templates using batch gradient descent optimization.
This training strategy was used to minimize the loss function $\mathcal{L}({\bm\theta})$ in Eq.~(\ref{eq:dcganappobj}), which is an approximation of the loss function in Eq.~(\ref{eq:idealobj}).

The face template extractor we used is based on FaceNet~\cite{Schroff_2015_CVPR}, one of the most accurate CNN models for face recognition currently available. 
To ensure that the face reconstruction scenario is realistic, we used an open-source implementation\footnote{\url{https://github.com/davidsandberg/facenet}} based on TensorFlow\footnote{Version 1.4.0 from https://www.tensorflow.org} without any modifications (model \textit{20170512-110547}). 

We implemented the \textit{NbNets} using MXNet\footnote{Version 0.1.0 from \url{http://mxnet.io}}.
The networks were trained using a mini-batch based algorithm, Adam~\cite{kingma2014adam} with batch size of 64, $\beta_1=0.5$ and $\beta_2=0.999$. 
The learning rate was initialized to $2\times10^{-4}$ and decayed by a factor of 0.94 every 5K batches. 
The pixel values in the output images were normalized to $[-1,1]$ by first dividing 127.5 and then subtracting 1.0. 
For the networks trained with the pixel difference loss, we trained the network with 300K batches, where the weights are randomly initialized using a normal distribution with zero mean and a standard deviation of 0.02.
For the networks trained with the perceptual loss~\cite{johnson2016perceptual}, we trained the networks with extra 100K batches by refining from the corresponding networks trained with the pixel difference loss. 
The hardware specifications of the workstations for the training are the CPUs of dual Intel(R) Xeon E5-2630v4 @ 2.2 GHz, the RAM of 256GB with two sets of NVIDIA Tesla K80 Dual GPU. 
The software includes CentOS 7 and Anaconda2\footnote{https://www.anaconda.com}. 

\begin{table*}[t]
	\caption{Deep template face template reconstruction models for comparison}
	\centering
	\setlength{\extrarowheight}{.4mm}
	\begin{threeparttable}
		\begin{tabular}{c|c|M{0.07\linewidth}|M{0.09\linewidth}|c}
			\hline
			Model\tnote{a}&Training Dataset & Training Loss&Testing Dataset&Training and Testing Process\\\hline \hline 
			VGG-Dn-P&\multirow{8}{*}{VGG-Face}&\multirow{3}{\linewidth}{Perceptual\\  Loss}&\multirow{21}{\linewidth}{LFW\\ FRGC v2.0\\ color FERET}&\multirow{7}{.47\linewidth}{Train a DCGAN using the training dataset, and then use face images generated from the pretrained DCGAN for training the target D-CNN. Test the trained D-CNN using testing datasets.}\\
			VGG-NbA-P&&&&\\
			VGG-NbB-P&&&&\\
			\cline{1-1}\cline{3-3}
			VGG-Dn-M&&\multirow{5}{\linewidth}{Pixel Difference (MAE\tnote{b} )}&&\\
			VGG-NbA-M&&&&\\
			VGG-NbB-M&&&&\\
			\cline{1-1}\cline{5-5}
			\multirow{2}{*}{VGGr-NbB-M}&&&&\multirow{2}{.47\linewidth}{Directly train the target D-CNN using face images from the training dataset, and then test the trained D-CNN using testing datasets.}\\
			&&&&\\
			\cline{1-3}\cline{5-5}
			MPIE-Dn-P&\multirow{8}{*}{Multi-PIE}&\multirow{3}{\linewidth}{Perceptual Loss}&&\multirow{7}{.47\linewidth}{Train a DCGAN using the training dataset, and then use face images generated from the pretrained DCGAN for training the target D-CNN. Test the trained D-CNN using testing datasets.}\\
			MPIE-NbA-P&&&&\\
			MPIE-NbB-P&&&&\\
			\cline{1-1}\cline{3-3}
			MPIE-Dn-M&&\multirow{8}{\linewidth}{Pixel Difference (MAE)}&&\\
			MPIE-NbA-M&&&&\\
			MPIE-NbB-M&&&&\\
			\cline{1-1}\cline{5-5}
			MPIEr-NbB-M&&&&\multirow{4}{.47\linewidth}{Directly train the target D-CNN using face images from the training dataset, and then test the trained D-CNN using testing datasets.}\\
			\cline{1-2}
			\multirow{3}{*}{Mixedr-NbB-M}&\multirow{3}{.12\linewidth}{VGG-Face CASIA-Webface  Multi-PIE}&&&\\
			&&&&\\
			&&&&\\
			\hline
			\multirow{2}{*}{RBF\cite{mignon2013reconstructing}}&LFW&N/A&LFW&\multirow{2}{.47\linewidth}{Train and test the RBF regression based method using the training and testing images specified in the evaluation protocol.}\\
			\cline{2-4}
			&FRGC v2.0&N/A&FRGC v2.0&\\\hline
		\end{tabular}
		\begin{tablenotes}
			\item[a] Dn, NbA, and NbB denote D-CNN, NbNet-A, and NbNet-B, respectively
			\item[b] MAE denotes ``mean absolute error''
		\end{tablenotes}
		\label{tb:compared_methods}
	\end{threeparttable}
\end{table*}

\section{Experimental Results}
\label{sec:experiment}

\subsection{Database and Experimental Setting}
The vulnerabilities of deep templates under template reconstruction attacks were studied with our proposed reconstruction model, using two popular large-scale face datasets for training and three benchmark datasets for testing. 
The training datasets consisted of one unconstrained datasets, VGG-Face~\cite{Parkhi15} and one constrained dataset, Multi-PIE~\cite{gross2007cmu}. 
\begin{itemize}
  \item \textbf{VGG-Face}~\cite{Parkhi15} comprises of 2.6 million face images from 2,622 subjects. 
  In total, 1,942,242 trainable images were downloaded with the provided links.
  \item \textbf{Multi-PIE}~\cite{gross2007cmu}. We used 150,760 frontal images (3 camera views, with labels $`14\_0', `05\_0'$, and $`05\_1'$, respectively), from 337 subjects.
\end{itemize}

Three testing datasets were used, including two for verification (LFW~\cite{learned2016labeled} and FRGC v2.0~\cite{phillips2005overview}) and one for identification (color FERET \cite{phillips2000feret}) scenarios.
\begin{itemize}
  \item \textbf{LFW}~\cite{learned2016labeled} consists of 13,233 images of 5,749 subjects downloaded from the Web.
  \item \textbf{FRGC v2.0}~\cite{phillips2005overview} consists of 50,000 frontal images of 4,003 subjects with two different facial expressions (smiling and neutral), taken under different illumination conditions.
  A total of 16,028 images of 466 subjects (as specified in the target set of Experiment 1 of FRGC v2.0~\cite{phillips2005overview}) were used. 
  \item \textbf{Color FERET}~\cite{phillips2000feret} consists of four partitions (i.e., \textit{fa, fb, dup1} and \textit{dup2}), including 2,950 images. Compared to the gallery set \textit{fa}, the probe sets (\textit{fb, dup1} and \textit{dup2}) contain face images of difference facial expression and aging.
\end{itemize}

\begin{figure*}[!t]
	\centering
	\subfloat[LFW]{{\includegraphics[width=.89\linewidth]{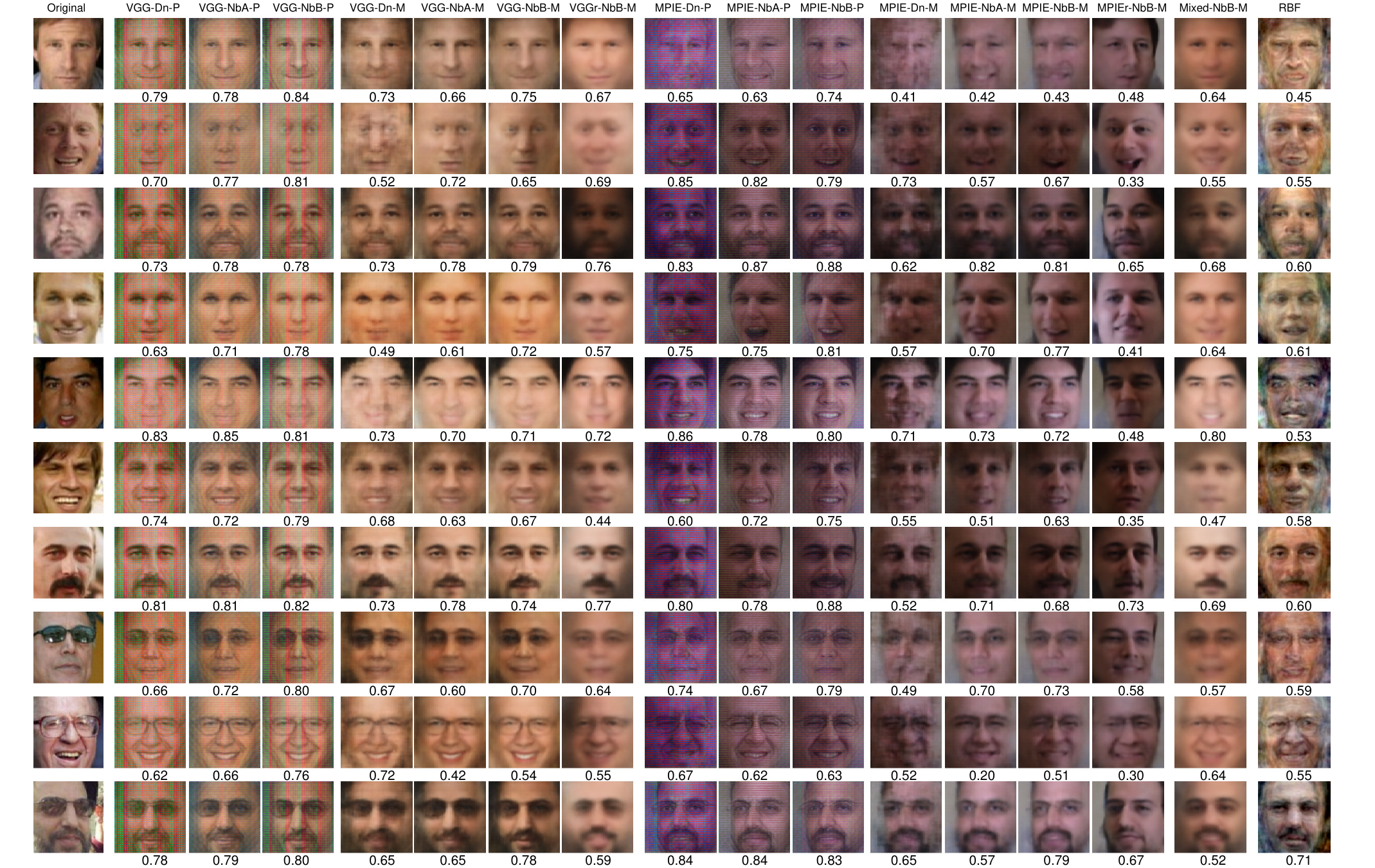}%
			\label{}}}\\
	\subfloat[FRGC v2.0]{{\includegraphics[width=.89\linewidth]{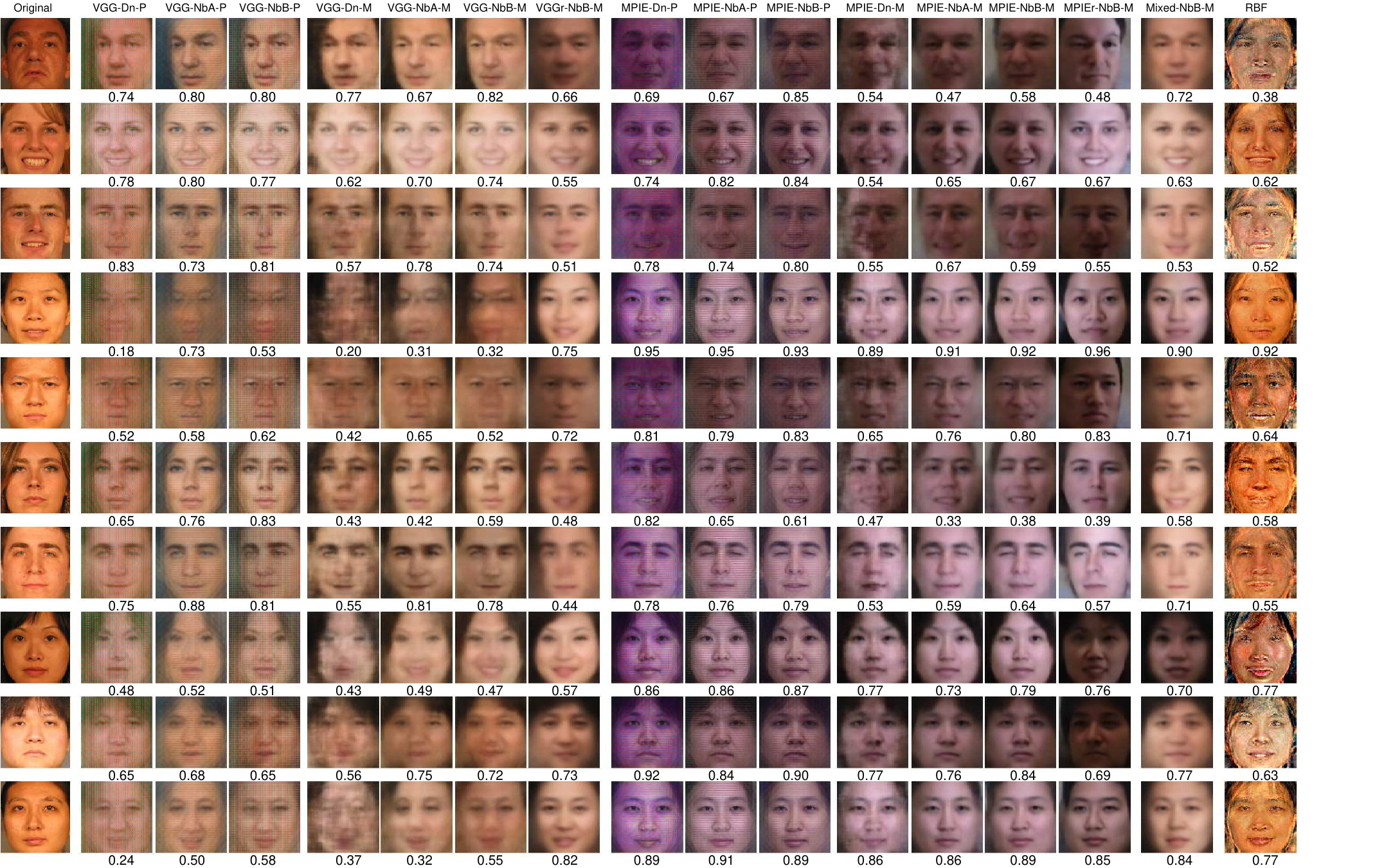}%
			\label{}}}\\
	\caption{Reconstructed face images of the first 10 subjects from (a) LFW and (b) FRGC v2.0. 
		The original face images are shown in the first column. Each column denotes the reconstructed face images from different models used for reconstruction.  
		The number below each reconstructed image shows the similarity score between the reconstructed image and the original image.
		The scores (ranging from -1 to 1) were calculated using the cosine similarity. 
		The mean verification thresholds on LFW and FRGC v2.0 were 0.51 and 0.80, respectively, at FAR=0.1\%, and 0.38 and 0.64, respectively, at FAR=1.0\%.}
	\label{fig:reconstruct_example}
\end{figure*}

The face images were aligned using the five points detected by MTCNN\footnote{\url{https://github.com/pangyupo/mxnet_mtcnn_face_detection.git}} \cite{zhang2016joint} and then cropped to $160\times160$ pixels. 
Instead of aligning images from the LFW dataset, we used the pre-aligned deep funneled version~\cite{Huang2012a}. 
Fig.~\ref{fig:datasetexample} shows example images from these five datasets.

\begin{figure*}[t]
	\centering
	\includegraphics[width=.75\linewidth]{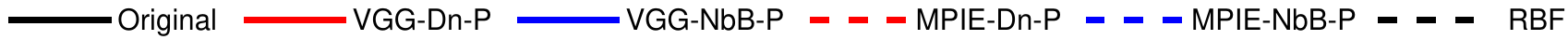}
	\subfloat[Type-I attack]{\includegraphics[width=.38\linewidth]{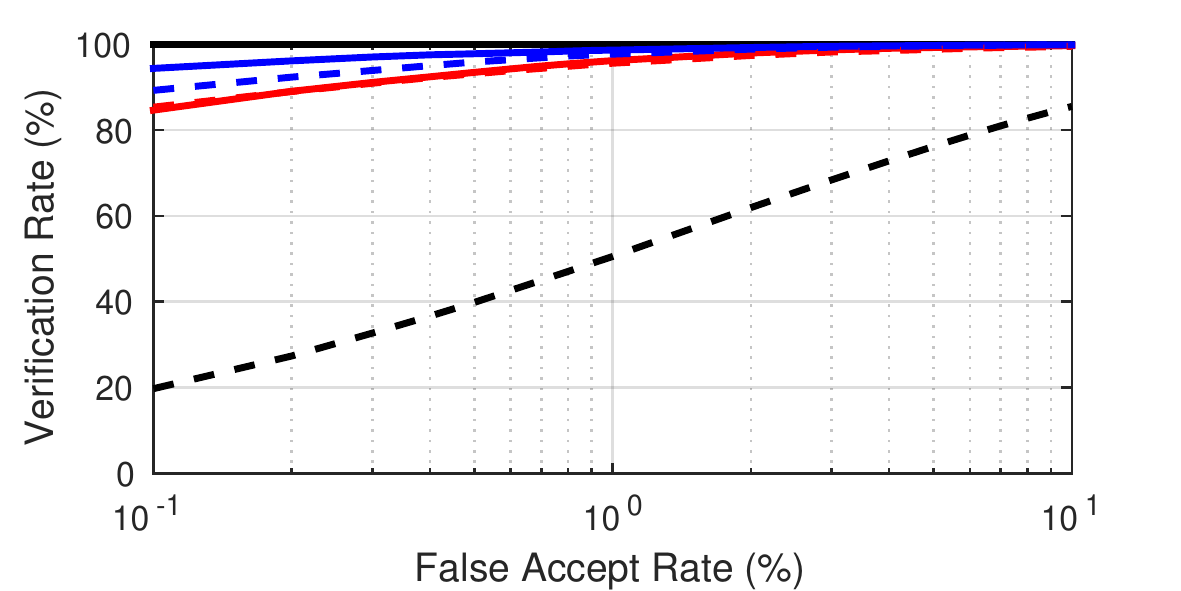}%
		\label{fig:type1lfw}} \hspace{0.5cm}
	\subfloat[Type-II attack]{\includegraphics[width=.38\linewidth]{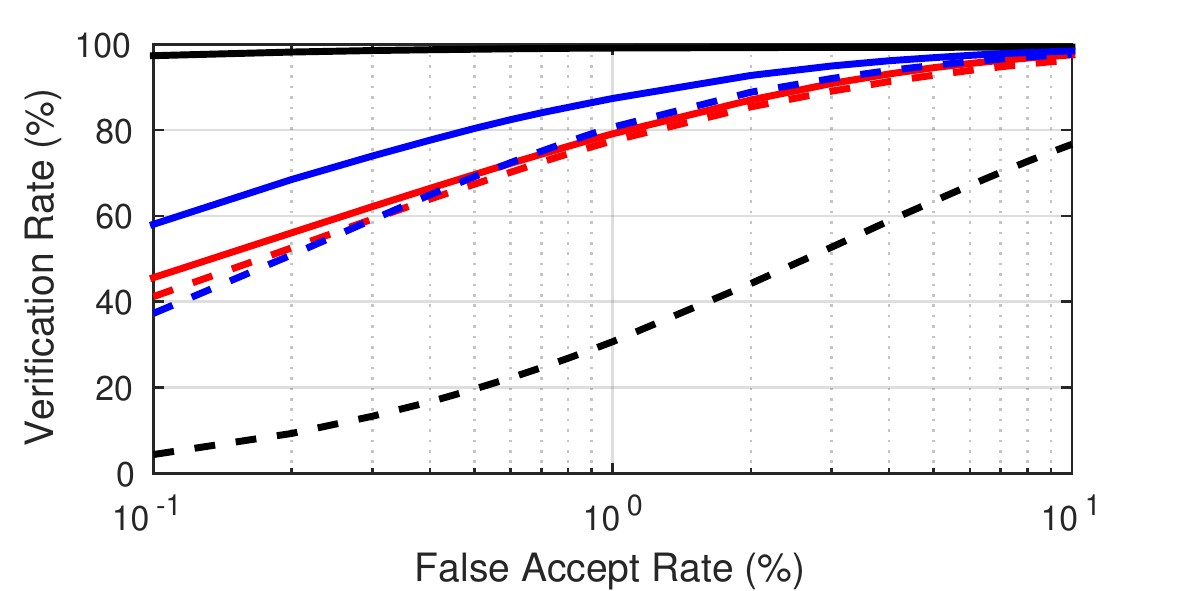}%
		\label{fig:type2lfw}}
	\caption{ROC curves of (a) type-I and (b) type-II attacks using different reconstruction models on LFW. For the ease of reading, we only show the curves for D-CNN, NbNet-B trained with perceptual loss, and the RBF based method. Refer to Table~\ref{tb:type1lfw} for the numerical comparison of all models. Note that the curves for VGG-Dn-P and MPIE-Dn-P are overlapping in (a).}
	\label{fig:roc_lfw}
\end{figure*}

To determine the effectiveness of the proposed \textit{NbNet}, we compare three different network architectures, i.e., D-CNN, NbNet-A, and NbNet-B, which are built using the typical de-convolution blocks, NbBlocks A and B. 
All of these networks are trained using the proposed generator-based training strategy using a DCGAN~\cite{radford2015unsupervised} with both pixel difference\footnote{We simply choose mean absolute error (MAE), where order $k=1$.} and perceptual loss\footnote{To reduce the training time, we first train the network with pixel difference loss and then fine-tune it using perceptual loss \cite{johnson2016perceptual}.} \cite{johnson2016perceptual}. 
To demonstrate the effectiveness of the proposed training strategy, we train the NbNet-B directly using images from VGG-Face, Multi-PIE, and a mixture of three datasets (VGG-Face, CASIA-Webface\footnote{It consists of 494,414 face images from 10,575 subjects. 
We obtain 455,594 trainable images after preprocessing.} \cite{yi2014learning}, and Multi-PIE). 
Note that both VGG-Face and Multi-PIE are augmented to 19.2M images in our training. 
Examples of images generated using our trained face image generator are shown in Fig.~\ref{fig:gen_example}.
In addition, the proposed \textit{NbNet} based reconstruction method was compared with a state-of-the-art RBF-regression-based method~\cite{mignon2013reconstructing}.
In contrast to the neural network based method, the RBF\footnote{It was not compared in the identification task on color FERET.} regression model of~\cite{mignon2013reconstructing} used the same dataset for training and testing (either LFW or FRGC v2.0).
Therefore, the RBF-regression-based reconstruction method was expected to have better reconstruction accuracy than the proposed method.
The MDS-based method~\cite{mohanty2007scores} was not compared here because it is a linear model and was not as good as the RBF-regression-based method~\cite{mignon2013reconstructing}.
We did not compare~\cite{zhmoginov2016inverting,cole2017synthesizing} because \cite{zhmoginov2016inverting} does not satisfy our assumption of \textit{black-box}  template extractor and \cite{cole2017synthesizing} requires to selecting high quality images for training. 
Table~\ref{tb:compared_methods} summarizes the 16 comparison models used in this study for deep template inversion.

Examples of the reconstructed images of the first ten subjects in LFW and FRGC v2.0 are shown in Fig.~\ref{fig:reconstruct_example}. 
The leftmost column shows the original images, and the remaining columns show the images reconstructed using the 16 reconstruction models.
For the RBF model, every image in the testing datasets (LFW and FRGC v2.0) has 10 different reconstructed images that can be created using the 10 cross-validation trials in the BLUFR protocol\footnote{\url{http://www.cbsr.ia.ac.cn/users/scliao/projects/blufr/}}~\cite{liao2014benchmark}.
The RBF-reconstructed images shown in this paper are those with the highest similarity scores among these 10 different reconstructions. 
The number below each image is the similarity score between the original and reconstructed images.
The similarity scores were calculated using the cosine similarity in the range of $[-1,1]$. 

\subsection{Verification Under Template Reconstruction Attack}
\label{sec:verification}

We quantitatively evaluated the template security of the target face recognition system (FaceNet \cite{Schroff_2015_CVPR}) under type-I and type-II template reconstruction attacks. 
The evaluation metric was face verification using the BLUFR protocol~\cite{liao2014benchmark}.
The impostor scores obtained from the original face images were used in both of the attacks to demonstrate the efficacy of the reconstructed face images.
The genuine scores in the type-I attack were obtained by comparing the reconstructed images against the original images.
The genuine scores in the type-II attack were obtained by substituting one of the original images in a genuine comparison (image pair) with the corresponding reconstructed image.
For benchmarking, we report the ``Original'' results based on original face images.
Every genuine score of ``Original'' in type-I attack was obtained by comparing two identical original images and thus the corresponding TAR stays at 100\%.
The genuine scores of ``Original'' in type-II attack were obtained by the genuine comparisons specified in BLUFR protocol.
The BLUFR protocol uses tenfold cross-validation; the performance reported here is the `lowest', namely $(\mu - \sigma)$, where $\mu$ and $\sigma$ denote the mean and standard deviation of the attacking accuracy obtained from the 10 trials, respectively.

\begin{table}
  \caption{TARs (\%) of type-I and type-II attacks on LFW for different template reconstruction methods, where ``Original'' denotes results based on the original images and other methods are described in Table~\ref{tb:compared_methods}. (\textbf{best}, \underline{second best})}
  \centering
  \setlength{\extrarowheight}{.4mm}
  \begin{tabular}{|c||c|c||c|c|}
        \hline
    Attack&\multicolumn{2}{c||}{Type-I}&\multicolumn{2}{c|}{Type-II}\\\hline
        FAR&0.1\%    &1.0\% &0.1\%   &1.0\% \\\hline 
        Original    &100.00  &100.00    &97.33    &99.11    \\\hline\hline 
        VGG-Dn-P     &84.65    &96.18    &45.63    &79.13   \\\hline
        VGG-NbA-P    &\textbf{95.20}    &\textbf{99.14}    &\underline{53.91}    &\underline{87.06}   \\\hline
        VGG-NbB-P    &\underline{94.37}    &\underline{98.63}   & \textbf{58.05}    &\textbf{87.37}   \\\hline\hline 
        VGG-Dn-M     &70.22    &88.35   &26.22    &64.88    \\\hline
        VGG-NbA-M    &79.52    &94.94    &30.97    &68.14   \\\hline
        VGG-NbB-M    &89.52    &97.75    &37.09    &79.19   \\\hline\hline 
        VGGr-NbB-M   &72.53    &93.21     &27.38    &70.72  \\\hline\hline 
        MPIE-Dn-P    &85.34    &95.57     &41.21    &77.51  \\\hline
        MPIE-NbA-P   &80.33    &95.46     &21.75    &63.05  \\\hline
        MPIE-NbB-P   &89.25    &97.69     &37.30    &80.67  \\\hline\hline 
        MPIE-Dn-M    &37.11    &63.01    &3.23    &13.26    \\\hline
        MPIE-NbA-M   &50.54    &78.91     &6.11    &33.26   \\\hline
        MPIE-NbB-M   &67.86    &88.56     &24.00    &57.98  \\\hline\hline 
        MPIEr-NbB-M  &34.87    &65.56      &3.67    &21.24  \\\hline\hline 
        Mixedr-NbB-M  &71.62    &92.98      &19.29    &65.63 \\\hline\hline 
        RBF~\cite{mignon2013reconstructing}      &19.76    &50.55   &4.41    &30.70    \\\hline
  
  \end{tabular}
  \label{tb:type1lfw}
\end{table}

\subsubsection{Performance on LFW}

In each trial of the BLUFR protocol~\cite{liao2014benchmark} for LFW~\cite{learned2016labeled}, there is an average of 46,960,863 impostor comparisons.
The average number of testing images is 9,708. Hence, there are 9,708 genuine comparisons in a type-I attack on LFW.
The average number of genuine comparisons in a type-II attack on LFW is 156,915; this is the average number of genuine comparisons specified in the BLUFR protocol.

The receiver operator characteristic (ROC) curves of type-I and type-II attacks on LFW are shown in Fig.~\ref{fig:roc_lfw}.
Table~\ref{tb:type1lfw} shows the TAR values at FAR=0.1\% and FAR=1.0\%, respectively.
The ROC curve of ``Original'' in the type-II attack (Fig.~\ref{fig:type2lfw}) is the system performance with BLUFR protocol~\cite{liao2014benchmark} based on original images.

For both type-I and type-II attacks, the proposed \textit{NbNets} generally outperform the D-CNN, where MPIE-NbA-P is not as effective as MPIE-Dn-P. 
Moreover, the models trained using the proposed strategy (VGG-NbB-M and MPIE-NbB-M) outperform the corresponding models trained with the non-augmented datasets (VGGr-NbB-M and MPIEr-NbB-M). 
The models trained using the raw images in VGG (VGGr-NbB-M) outperform the corresponding model trained using the mixed dataset. 
All \textit{NbNets} trained with the proposed training strategy outperform the RBF regression based method \cite{mignon2013reconstructing}.
In the type-I attack, the VGG-NbA-P model achieved a TAR of 95.20\% (99.14\%) at FAR=0.1\% (FAR=1.0\%).  
This implies that an attacker has approximately 95\% (or 99\% at FAR=1.0\%) chance of accessing the system using a leaked template.

\begin{figure*}[!t]
	\centering
	\includegraphics[width=.75\linewidth]{figure/pdf/legend.pdf}
	\subfloat[Type-I attack]{\includegraphics[width=.38\linewidth]{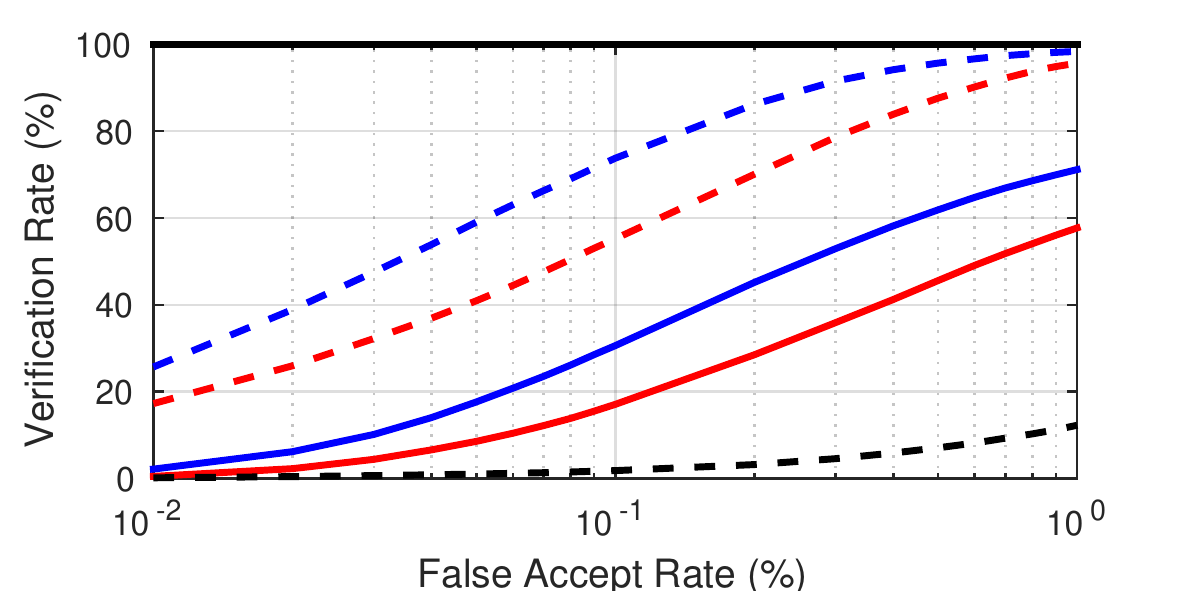}%
		\label{fig:type1frgc}}  \hspace{0.5cm}
	\subfloat[Type-II attack]{\includegraphics[width=.38\linewidth]{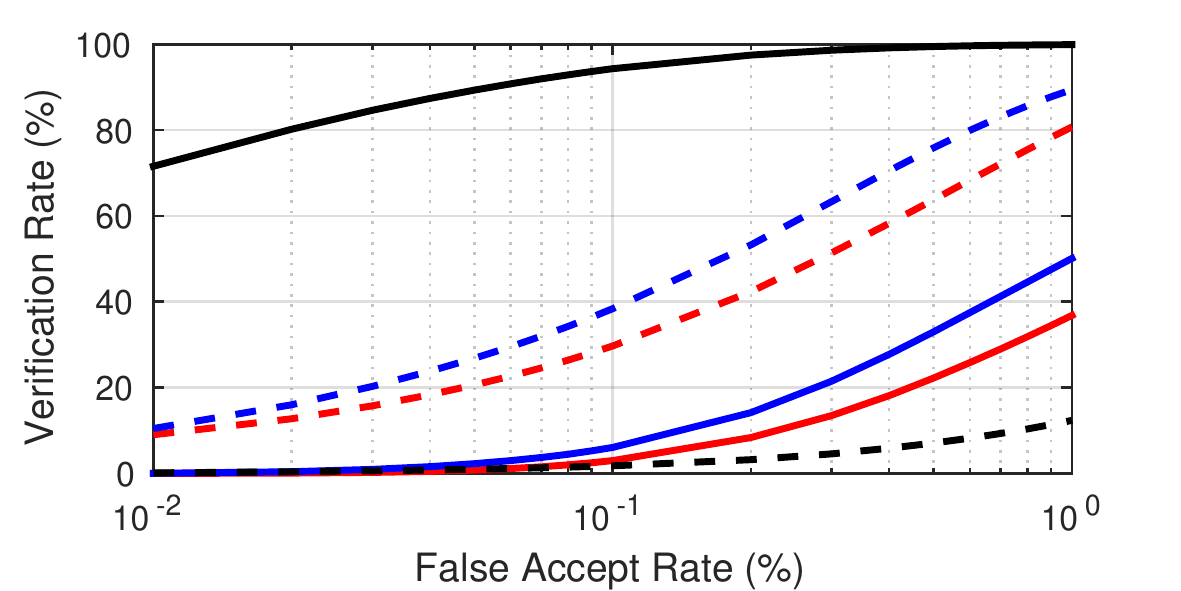}%
		\label{fig:type2frgc}}
	\caption{ROC curves of (a) type-I and (b) type-II attacks using different reconstruction models  on FRGC v2.0. For readability, we only show the curves for D-CNN, NbNet-B trained with perceptual loss, and the RBF based method. Refer to Table~\ref{tb:type1frgc} for the numerical comparison of all models.}
	\label{fig:roc_frgc}
\end{figure*}

\subsubsection{Performance on FRGC v2.0}

Each trial of the BLUFR protocol~\cite{liao2014benchmark} for FRGC v2.0~\cite{phillips2005overview} consisted of an average of 76,368,176 impostor comparisons and an average of 12,384 and 307,360 genuine comparisons for type-I and type-II attacks, respectively.

\begin{table}
  \caption{TARs (\%) of type-I and type-II attacks on FRGC v2.0 for different template reconstruction methods, where ``Original'' denotes results based on the original images and other methods are described in Table~\ref{tb:compared_methods}. (\textbf{best}, \underline{second best}) }
  \centering
  \setlength{\extrarowheight}{.4mm}
  \begin{tabular}{|c||c|c||c|c|}
    \hline
    Attack&\multicolumn{2}{c||}{Type-I}&\multicolumn{2}{c|}{Type-II}\\\hline
    FAR&0.1\%    &1.0\% &0.1\%   &1.0\% \\\hline 
      Original     &100.00   &100.00  &94.30    &99.90    \\\hline\hline 
      VGG-Dn-P      &17.10    &57.71    &3.00    &36.81    \\\hline
      VGG-NbA-P     &32.66    &71.54    &8.65    &51.87    \\\hline
      VGG-NbB-P     &30.62    &71.14    &6.06    &50.09    \\\hline\hline
      VGG-Dn-M      &3.52     &35.94   &0.68    &20.40     \\\hline
      VGG-NbA-M     &8.95     &55.84    &2.39    &33.40    \\\hline
      VGG-NbB-M     &16.44    &67.57    &3.60    &44.19    \\\hline\hline
      VGGr-NbB-M      &6.75     &55.51     &4.05    &36.18   \\\hline\hline
      MPIE-Dn-P       &\underline{55.22}    &\underline{95.65}     &\underline{29.70}    &\underline{80.72}  \\\hline
      MPIE-NbA-P      &49.75    &94.41     &28.46    &78.71  \\\hline
      MPIE-NbB-P      &\textbf{73.76}    &\textbf{98.35}     &\textbf{38.39}    &\textbf{89.41}  \\\hline\hline
      MPIE-Dn-M     &12.82    &47.84    &10.47    &38.39   \\\hline
      MPIE-NbA-M      &15.58    &61.44     &13.42    &48.46  \\\hline
      MPIE-NbB-M      &28.48    &80.67     &19.85    &63.04  \\\hline\hline
      MPIEr-NbB-M     &12.72    &49.53      &11.75    &40.59 \\\hline\hline
      Mixedr-NbB-M     &9.65     &63.82      &8.15    &45.10  \\\hline\hline
      RBF~\cite{mignon2013reconstructing}      &1.86     &12.29   &1.78    &12.37    \\\hline
  \end{tabular}
  \label{tb:type1frgc}
\end{table}

The ROC curves of type-I and type-II attacks on FRGC v2.0 are shown in Fig.~\ref{fig:roc_frgc}. 
The TAR values at FAR=0.1\% and FAR=1.0\% are shown in Table~\ref{tb:type1frgc}.
The TAR values (Tables \ref{tb:type1lfw} and \ref{tb:type1frgc}) and ROC plots (Figs. \ref{fig:roc_lfw} and \ref{fig:roc_frgc}) for LFW and FRGC v2.0 cannot be directly compared, as the thresholds for LFW and FRGC v2.0 differ (e.g., the thresholds at FAR=0.1\% are 0.51 and 0.80 for LFW and FRGC v2.0, respectively). 
The similarity threshold values were calculated based on the impostor distributions of the LFW and FRGC v2.0 databases.

It was observed that the proposed \textit{NbNets} generally outperform D-CNN. 
The only exception is that the MPIE-NbA-P was not as good as MPIE-Dn-P. 
Significant improvements by using the augmented datasets (VGG-NbB-M and MPIE-NbB-M) were observed, compared with VGGr-NbB-M and MPIEr-NbB-M, for both the type-I and type-II attacks.
All \textit{NbNets} outperform the RBF regression based method \cite{mignon2013reconstructing}.
In the type-I attack, the best model, MPIE-NbB-P achieved a TAR of 73.76\% (98.35\%) at FAR=0.1\% (FAR=1.0\%). 
This implies that an attacker has a 74\% (98\%) chance of accessing the system at FAR=0.1\% (1.0\%) using a leaked template.

\begin{table}
  \caption{Rank-one recognition rate (\%) on color FERET \cite{phillips2000feret} with partition \textit{fa} as gallery and reconstructed images from different partition as probe.
  	The partitions (i.e., \textit{fa}, \textit{fb}, \textit{dup1} and \textit{dup2}) are described in color FERET protocol  \cite{phillips2000feret}.
  Various methods are described in Table~\ref{tb:compared_methods}. (\textbf{best} and \underline{second best}) of rank-one identification rate in each column.}
  \centering
  \setlength{\extrarowheight}{.4mm}
  \begin{tabular}{|c||c||c|c|c|}
        \hline
        Attack       &Type-I   &\multicolumn{3}{c|}{Type-II}\\\hline
        Probe        &fa       &fb    &dup1  &dup2 \\\hline 
        VGG-Dn-P     &89.03    &86.59 &76.77 &78.51\\\hline
        VGG-NbA-P    &94.87    &90.93 &80.30 &81.58\\\hline
        VGG-NbB-P    &95.57    &\textbf{92.84} &\underline{84.78} &84.65\\\hline\hline 
        VGG-Dn-M     &80.68    &74.40 &62.91 &65.35\\\hline
        VGG-NbA-M    &86.62    &80.44 &64.95 &66.67\\\hline
        VGG-NbB-M    &92.15    &87.00 &75    &75.44\\\hline\hline 
        VGGr-NbB-M   &81.09    &74.29 &61.28 &62.28\\\hline\hline 
        MPIE-Dn-P    &\underline{96.07}    &91.73 &84.38 &\underline{85.53}\\\hline
        MPIE-NbA-P   &93.86    &90.22 &79.89 &79.82\\\hline
        MPIE-NbB-P   &\textbf{96.58}    &\textbf{92.84} &\textbf{86.01} &\textbf{87.72}\\\hline\hline 
        MPIE-Dn-M    &73.54    &64.11 &53.26 &49.12\\\hline
        MPIE-NbA-M   &72.23    &64.01 &51.09 &44.74\\\hline
        MPIE-NbB-M   &85.61    &78.22 &71.06 &68.42\\\hline\hline 
        MPIEr-NbB-M  &63.88    &54.54 &44.57 &35.96\\\hline\hline 
        Mixedr-NbB-M &82.19    &76.11 &62.09 &58.77\\\hline\hline 
         Original    &100.00   &98.89 &97.96 &99.12\\\hline
  \end{tabular}
  \label{tb:idferet}
\end{table}

\subsection{Identification with Reconstructed Images}
\label{sec:exp_iden}

We quantitatively evaluate the privacy issue of a leaked template extracted by target face recognition system (FaceNet \cite{Schroff_2015_CVPR}) under type-I and type-II attacks. 
The evaluation metric was the standard color FERET protocol \cite{phillips2000feret}. 
The partition \textit{fa} (994 images)  was used as the gallery set.
For the type-I attack, the images reconstructed from the partition \textit{fa} was used as the probe set. 
For the type-II attack, the probe sets (\textit{fb} with 992 images, \textit{dup1} with 736 images, and \textit{dup2} with 228 images) specified in the color FERET protocol were replaced by the corresponding reconstructed images.  

The rank-one identification rate of both type-I and type-II attacks on color FERET are shown in Table~\ref{tb:idferet}. 
The row values under "Original" show the identification rate based on the original images. 
It stays at 100\% for the type-I attack because the corresponding similarity score are obtained by comparing two identical images. 
It was observed that the proposed \textit{NbNets} outperform D-CNN with the exception that the MPIE-Dn-P and MPIE-Dn-M slightly outperform MPIE-NbA-P and MPIE-NbA-M, respectively. 
Besides, significant improvements introduced by the proposed training strategy were observed, comparing models VGG-NbB-M and MPIE-NbB-M with the corresponding models trained with raw images (VGGr-NbB-M and MPIEr-NbB-M), respectively. 
It was observed that the best model, MPIE-NbB-P achieves 96.58\% and 92.84\% accuracy under type-I and type-II attacks (partition \textit{fb}). 
This implies a severe privacy issue; more than 90\% of the subjects in the database can be identified with a leaked template.

\subsection{Computation Time}
In the testing stage, with an NVIDIA TITAN X Pascal (GPU) and an Intel(R) i7-6800K @ 3.40 GHz (CPU), the average time (in microseconds) to reconstruct a single face template with D-CNN, NbNet-A, and NbNet-B is shown in Table~\ref{tb:comptime}.
\begin{table}
  \caption{Average reconstruction time (ms) for a single template. The total number of network parameters are indicated in the last column.}
  \centering
  \setlength{\extrarowheight}{.4mm}
  \begin{tabular}{|c|c|c|c|}
        \hline
               & CPU & GPU & \#Params\\ \hline
        D-CNN    & 84.1 & 0.268   & 4,432,304\\ \hline
        NbNet-A  & 62.6  & 0.258  & 2,289,040\\ \hline
        NbNet-B  & 137.1 & 0.477  & 3,411,472\\ \hline
  \end{tabular}
  \label{tb:comptime}
\end{table}

\section{Conclusions and Future Work}
\label{sec:conclusion}

We investigated the security and privacy of deep face templates by studying the reconstruction of face images via the inversion of their corresponding deep templates.
A \textit{NbNet} was trained for reconstructing face images from their corresponding deep templates and strategies for training generalizable and robust \textit{NbNets} were developed.
Experimental results indicated that the proposed NbNet-based reconstruction method outperformed RBF-regression-based face template reconstruction in terms of attack success rates.
We demonstrate that in verification scenario, TAR of 95.20\% (58.05\%) on LFW under type-I (type-II) attack @ FAR of 0.1\% can be achieved with our reconstruction model. 
Besides, 96.58\% (92.84\%) of the images reconstruction from templates of partition \textit{fa} (\textit{fb}) can be identified from partition \textit{fa} in color FERET \cite{phillips2000feret}.
This study revealed potential security and privacy issues resulting from template leakage in state-of-the-art face recognition systems, which are primarily based on deep templates. 

Our future research goals are two-fold: protecting the system from template reconstruction attacks and improving the proposed reconstruction. 
For the protection, we aim to design a template protection scheme~\cite{feng2010hybrid,nandakumar2015biometric,lim2016learning,mai2016binary} by introducing user-specific randomness into deep networks for extracting secure and discriminative deep templates.
Therefore, the extracted deep templates not only depend on the input images, but also subject-specific keys. 
To further enhance the system security, stronger anti-spoofing techniques~\cite{wen2015face,patel2016spoofdetect,siqiliu20163d,shao2017deep,liu2018learning} will also be sought. 
For the improvement of the reconstruction, we plan to (\textit{i}) devise a more effective reconstruction algorithm by designing a more effective \textit{NbNet} and considering the holistic contents in face images; and 
(\textit{ii}) study cross-system attacks using face images reconstructed from the templates of a given face recognition system to access a different face recognition system (different from the one used to generate he template). 



\ifCLASSOPTIONcompsoc
  \section*{Acknowledgments}
\else
\fi

This research was partially supported by a Hong Kong RGC grant (HKBU 12201414)  and the Madam Kwok Chung Bo Fun Graduate School Development Fund, HKBU.
The authors would like to thank Prof. Wei-Shi Zheng, Dr. Xiangyuan Lan and Miss Huiqi Deng for their helpful suggestions.


\ifCLASSOPTIONcaptionsoff
  \newpage
\fi



%

\begin{IEEEbiography}[{\includegraphics[width=1in,height=1.25in,clip,keepaspectratio]{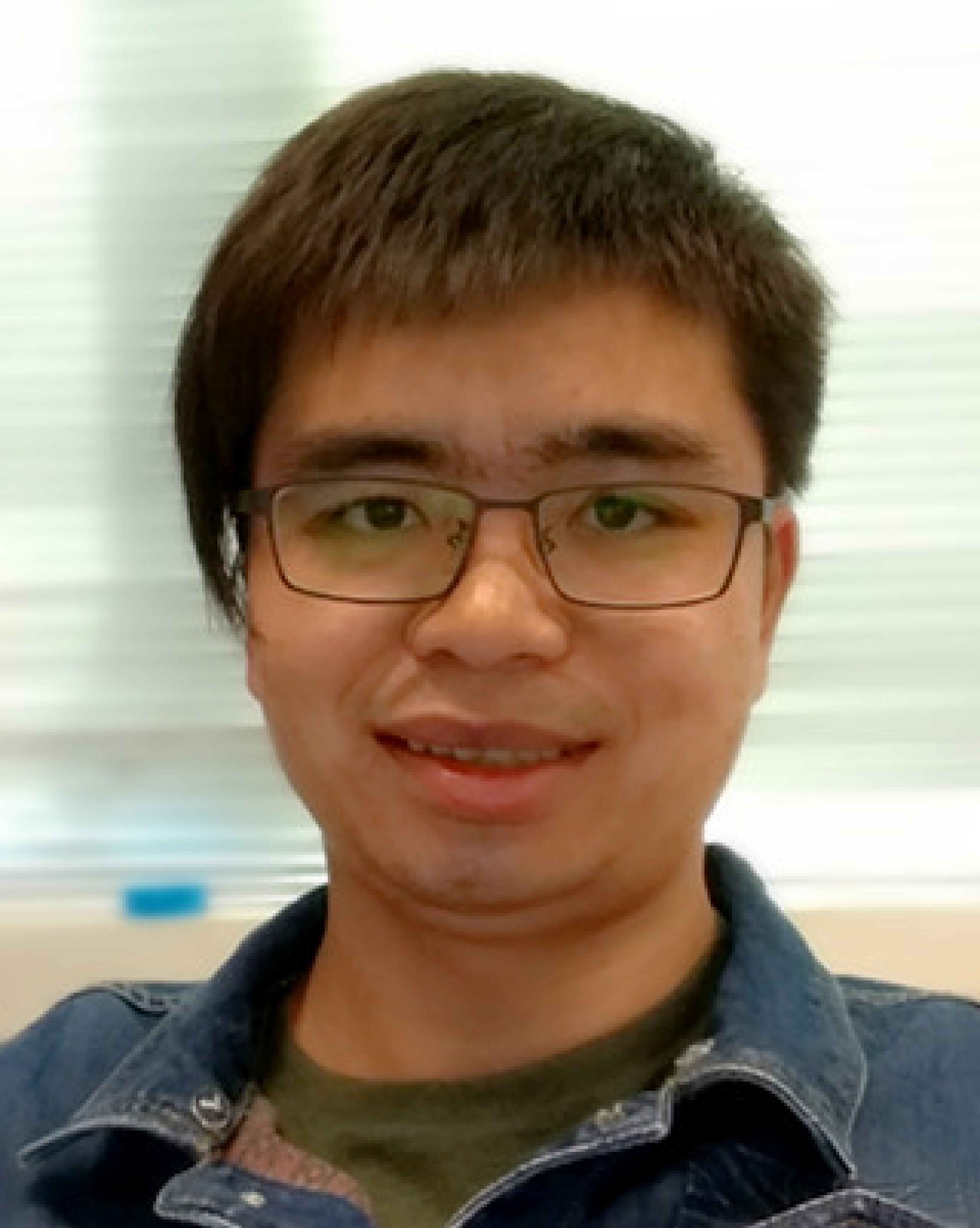}}]{Guangcan Mai}
received the B.Eng degree in computer science and technology from South China University of Technology, Guangzhou, China, in 2013. He is currently pursuing the Ph.D degree in Computer Science from Hong Kong Baptist University, Hong Kong. He was a Visiting Scholar at Michigan State University, USA. His research interests include biometric security and machine learning. He is a student member of IEEE.
\end{IEEEbiography}

\begin{IEEEbiography}[{\includegraphics[width=1in,height=1.25in,clip,keepaspectratio]{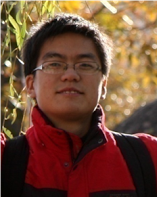}}]{Kai Cao}
	received the Ph.D. degree from Key Laboratory of Complex Systems and Intelligence Science, Institute of Automation, Chinese Academy of Sciences, Beijing, China, in 2010. He is currently a Post Doctoral Fellow in the Department of Computer
	Science and Engineering, Michigan State University, East Lansing. His research interests include biometric recognition, image processing and machine learning
\end{IEEEbiography}

\begin{IEEEbiography}[{\includegraphics[width=1in,height=1.25in,clip,keepaspectratio]{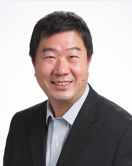}}]{Pong C. Yuen}
is a Professor of the Department of Computer Science in Hong Kong Baptist University.
He was the head of the department from 2009 to 2017. 
He is the Vice President (Technical Activities) of the IEEE Biometrics Council, Editorial Board Member of Pattern Recognition, Associate Editor of IEEE Transactions on Information Forensics and Security and Senior Editor of SPIE Journal of Electronic Imaging.  
He also serves as a Hong Kong Research Grant Council Engineering Panel Member. 
His current research interests include video surveillance, human face recognition, biometric security and privacy.
\end{IEEEbiography}

\begin{IEEEbiography}[{\includegraphics[width=1in,height=1.25in,clip,keepaspectratio]{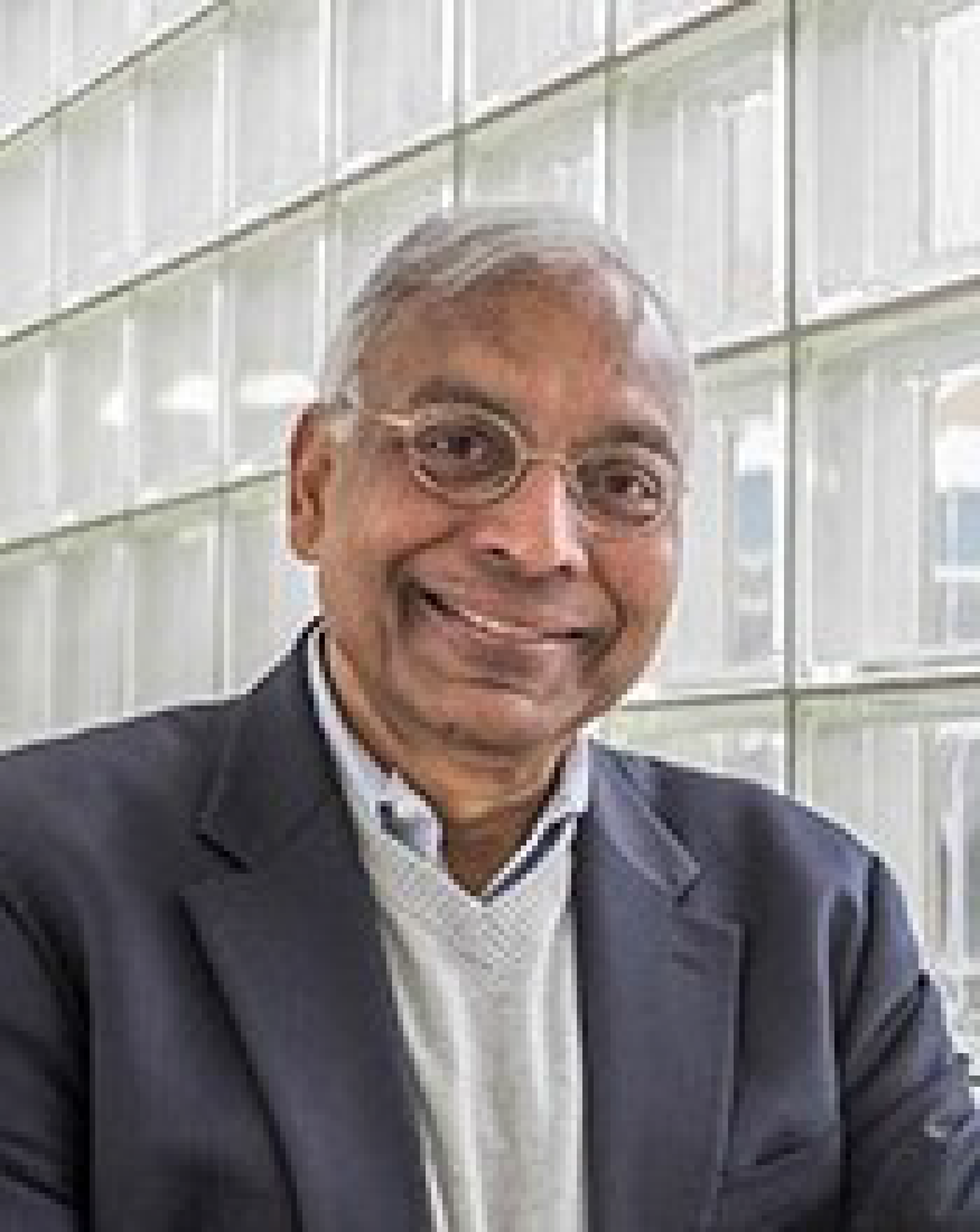}}]{Anil K. Jain}
is a University distinguished professor in the Department of Computer Science and Engineering at Michigan State University, East Lansing, Michigan.
He served as the editor-in-chief of the IEEE Transactions on Pattern Analysis and Machine Intelligence (1991-1994), a member of the United States Defense Science Board and a member of the Forensic Science Standards Board.
He has received Fulbright, Guggenheim, Alexander von Humboldt, and IAPR King Sun Fu awards. He is a member of the United States National Academy of Engineering and a Foreign Fellow of the Indian National Academy of Engineering.
\end{IEEEbiography}

\begin{appendices}
\onecolumn
\section{Proof of the Existence of a Face Image Generator}
\label{apdx:generator}
Suppose a face image $\bm{x}\in \mathbb{R}^{h\times w \times c}$ of height $h$, width $w$, and $c$ channels can be represented by a real vector $\bm{b}=\{b_1,\cdots,b_k\}\in \mathbb{R}^k$ in a manifold space with $h\times w\times c \gg k$, 
where $b_i\sim \mathcal{F}_{b_i}, i\in [1,k]$ and $\mathcal{F}_{b_i}$ is the cumulative distribution function of $b_i$.
The covariance matrix of $\bm{b}$ is $\Sigma_{\bm{b}}$.
Given a multivariate uniformly distributed random vector $z \in [0,1]^k$ consisting of $k$ independent variables, there exists a generator function $\bm{b'}=\hat{r}(\bm{z})$, $\bm{b'}=\{b'_1, \cdots, b'_k\}$ such that $b'_i\sim \mathcal{F}_{b_i}, i\in [1,k]$, and $\Sigma_{\bm{b'}}\cong \Sigma_{\bm{b}}$.

\begin{proof}
  The function $\hat{r}(\cdot)$ exists and can be constructed by first introducing an intermediate multivariate normal random vector $\bm{a}\sim \mathcal{N}(\bm{0},\Sigma_{\bm a})$, and then applying the following transformations:
  
  (a) NORTA~\cite{cario1997modeling,ghosh2003behavior}, which transforms vector $\bm a$ into vector $\bm b' = \{b'_1, \cdots, b'_k\}$ with  $b'_i\sim \mathcal{F}_{b_i}, i\in [1,k]$ and the corresponding covariance matrix $\Sigma_{\bm{b'}}\cong \Sigma_{\bm{b}}$ by adjusting the covariance matrix $\Sigma_{\bm{a}}$ of $\bm a$. 
  \begin{equation}
    b'_i = \mathcal{F}^{-1}_{b_i}\left[\Phi(a_i)\right],i\in [1,k],
  \end{equation}
  where $\Phi(\cdot)$ denotes the univariate standard normal cdf and $\mathcal{F}^{-1}_{b_i}(\cdot)$ denotes the inverse of $\mathcal{F}_{b_i}$. 
  To achieve $\Sigma_{\bm{b'}}\cong \Sigma_{\bm{b}}$, a matrix $\Lambda_{\bm a}$ that denotes the covariance of the input vector $\bm a$ can be uniquely determined~\cite{henderson2000generating}. 
  If $\Lambda_{\bm a}$ is a feasible covariance matrix (symmetric and positive semi-definite with unit diagonal elements; a necessary but insufficient condition), $\Sigma_{\bm a}$ can be set to $\Lambda_{\bm a}$. Otherwise, $\Sigma_{\bm a}$ can be approximated by solving the following equation:
  \begin{equation}
    \begin{split}
      &\arg \min_{\Sigma_{\bm a}} \mathcal{D}(\Sigma_{\bm a},\Lambda_{\bm a})\\
      &\text{subject to } \Sigma_{\bm a}\ge 0, \Sigma_{\bm a}(i,i)=1
    \end{split}
  \end{equation}
  where $\mathcal{D}(\cdot)$ is a distance function~\cite{ghosh2003behavior}.
  
  (b) Inverse transformation~\cite{johnson2013multivariate} to generate $\bm{a}\sim \mathcal{N}(\bm{0},\Sigma_{\bm a})$ from multivariate uniformly distributed random vector $\bm{z}=\{z_1,\cdots,z_k\}$, where $z_i\sim \mathbf{U}(0,1), i\in[1,k]$. 
  \begin{equation}
    \bm{a} = \mathbf{M}\cdot\left[\Phi^{-1}(z_1), \cdots, \Phi^{-1}(z_k)\right]'
  \end{equation}
  where $M$ is a lower-triangular, non-singular, factorization of $\Sigma_{\bm a}$ such that $\mathbf{M}\mathbf{M}' = \Sigma_{\bm a}$, $\Phi^{-1}$ is the inverse of the univariate standard normal cdf~\cite{johnson2013multivariate}.
  
  This completes the proof.
\end{proof}

\end{appendices}
\end{document}